\documentclass[10pt,twocolumn,letterpaper]{article}

\usepackage{cvpr}              % To produce the CAMERA-READY version
\usepackage{amsmath}
\usepackage{amssymb}
\usepackage{graphicx}
\usepackage{booktabs}
\usepackage{multirow}
\usepackage{tikz}
\usepackage{url}
\usepackage{xcolor}
\usepackage{graphicx}
\usepackage{adjustbox}
\usepackage{colortbl}
\usepackage{float}
\definecolor{denomOrange}{rgb}{1.0, 0.9, 0.8} % Pastel Orange
\definecolor{numPink}{rgb}{1.0, 0.85, 0.9}    % Pastel Pink
 
\newcommand{\mypar}[1]{\vspace{0.0em}\noindent\textbf{#1}\textbf{.}}
\usepackage{tcolorbox}
\definecolor{myorange}{HTML}{E69F00}
\definecolor{myblue}{HTML}{57B4E9}
\definecolor{mediumgrey}{HTML}{A0A0A0} % Defines a standard grey

\newcommand{\greycircle}{%
  \tikz[baseline=-0.6ex]{
    \draw[line width=0.6pt, draw=mediumgrey] (-0.5em,0) -- (0.5em,0);
    \filldraw[fill=mediumgrey, draw=mediumgrey] (0,0) circle (0.6ex);
  }%
}

\newcommand{\orangecircle}{%
  \tikz[baseline=-0.6ex]{
    \draw[line width=0.6pt, draw=myorange] (-0.5em,0) -- (0.5em,0);
    \filldraw[fill=myorange, draw=myorange] (0,0) circle (0.6ex);
  }%
}
\newcommand{\greyrect}{%
  \tikz[baseline=-0.6ex, scale=0.4]{
    \draw[dashed, line width=0.7pt, draw=mediumgrey] (-0.5,0) -- (0.7,0);
    \fill[mediumgrey] (-0.25,-0.25) rectangle (0.25,0.25);
}}

\newcommand{\bluerect}{%
  \tikz[baseline=-0.6ex, scale=0.4]{
    \draw[dashed, line width=0.7pt, draw=myblue] (-0.5,0) -- (0.7,0);
    \fill[myblue] (-0.25,-0.25) rectangle (0.25,0.25);
  }%
}

\newcommand{\bluecircle}{%
  \tikz[baseline=-0.6ex]{
    \draw[line width=0.6pt, draw=myblue] (-0.5em,0) -- (0.5em,0);
    \filldraw[fill=myblue, draw=myblue] (0,0) circle (0.6ex);
  }%
}

\newcommand{\orangerect}{%
  \tikz[baseline=-0.6ex, scale=0.4]{
    \draw[dashed, line width=0.7pt, myorange] (-0.5,0) -- (0.7,0);
    \fill[myorange] (-0.25,-0.25) rectangle (0.25,0.25);
  }%
}

\definecolor{cvprblue}{rgb}{0.21,0.49,0.74}
\usepackage[pagebackref,breaklinks,colorlinks,allcolors=cvprblue]{hyperref}

\usepackage{amssymb}
\usepackage{amsmath,amsfonts}
\usepackage{amsopn}
\usepackage{bm} %
\usepackage{multirow}
\usepackage{tabularx}

\newcommand{\mat}[1]{\boldsymbol{#1}} %

\newcommand{\field}[1]{\mathbb{#1}}

\newcommand{\R}{\field{R}} %

\newcommand{\ProbOpr}[1]{\mathbb{#1}}

\newcommand{\expect}[2]{%
\ifthenelse{\equal{#2}{}}{\ProbOpr{E}_{#1}}
{\ifthenelse{\equal{#1}{}}{\ProbOpr{E}\left[#2\right]}{\ProbOpr{E}_{#1}\left[#2\right]}}} %
\newcommand{\var}[2]{%
\ifthenelse{\equal{#2}{}}{\ProbOpr{VAR}_{#1}}
{\ifthenelse{\equal{#1}{}}{\ProbOpr{VAR}\left[#2\right]}{\ProbOpr{VAR}_{#1}\left[#2\right]}}} %

\newcommand{\mA}{\mat{A}}

\newcommand{\eat}[1]{}

\definecolor{egogreen}{RGB}{214,235,197}
\definecolor{refred}{RGB}{251,210,204}
\definecolor{LightCyan}{rgb}{0.88,1,1}
\definecolor{LightRed}{rgb}{1,0.94,0.94}
\definecolor{LightBlue}{rgb}{0.94,0.97,1}
\definecolor{scarlet}{RGB}{147,0,0}
\definecolor{citeblue}{RGB}{238,26,28}
\definecolor{Blue}{RGB}{51,10,154}

\definecolor{citecolor}{RGB}{50, 63, 138}
\definecolor{linkcolor}{RGB}{187,18,26}
\def\paperID{} % *** Enter the Paper ID here
\def\confName{CVPR}
\def\confYear{2026}

\title{Revisiting Model Stitching \adjustbox{valign=c, height=1.3ex, raise=0.5ex}{\includegraphics{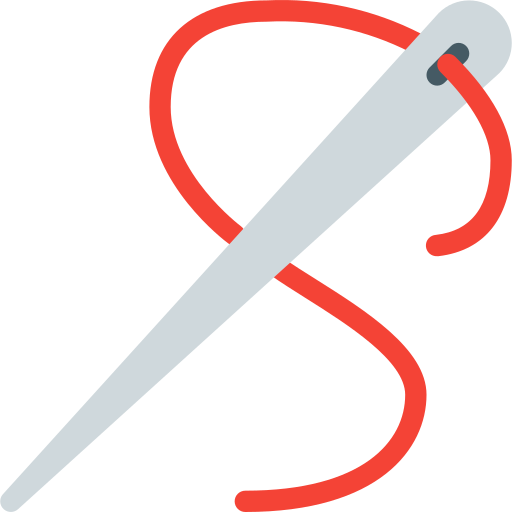}} in the Foundation Model Era}

\author{
Zheda Mai$^{1,3}$\thanks{Work done during internship at Amazon.}\hspace{0.6em} Ke Zhang$^{3}$\hspace{0.6em} Fu-En Wang$^{3}$\hspace{0.6em} Zixiao Ken Wang$^{3}$ \\Albert Y.~C.~Chen$^{3}$\hspace{0.6em}  Lu Xia$^{3}$\hspace{0.6em} Min Sun$^{3}$\hspace{0.6em} Wei-Lun Chao$^{2}$\hspace{0.6em} Cheng-Hao Kuo$^{3}$\\
$^{1}$The Ohio State University \qquad $^{2}$Boston University \qquad $^{3}$Amazon \\
{\tt\small \url{https://zheda-mai.github.io/Model-Stitch}}  \qquad  {\tt\small mai.145@osu.edu}
}
\begin{document}
\maketitle

% Abstract - Updated with new storyline emphasis
\begin{abstract}

Model stitching, connecting early layers of one model (source) to later layers of another (target) via a light stitch layer, has served as a probe of representational compatibility. Prior work finds that models trained on the same dataset remain stitchable (negligible accuracy drop) despite different initializations or objectives. We revisit stitching for Vision Foundation Models (VFMs) that vary in objectives, data, and modality (\eg, CLIP, DINOv2, SigLIP2) and ask: \emph{\textbf{Are heterogeneous VFMs stitchable?}} We introduce a systematic protocol spanning stitch positions, stitch layer families, training losses, and downstream tasks. Three findings emerge. (1) Stitch layer training matters: conventional approaches that match the intermediate features at the stitch position or optimize the task loss end-to-end struggle to retain accuracy, especially at shallow stitch positions. (2) With a simple feature-matching loss at the target model's penultimate layer, heterogeneous VFMs become reliably stitchable across vision tasks. (3) For deep stitch positions, the stitched model can significantly \textbf{surpass} either constituent model with a small inference overhead (for the stitch layer). Building on these findings, we further propose the VFM Stitch Tree (VST), which shares early layers across VFMs while retaining their later layers, yielding a controllable accuracy-latency trade-off for multimodal LLMs that often leverage multiple VFMs. Taken together, our study elevates stitching from a diagnostic probe to a practical recipe for integrating complementary VFM strengths and pinpointing where their representations align or diverge.
\end{abstract}

\section{Introduction}
\begin{figure}
    \centering
    \includegraphics[width=1\linewidth]{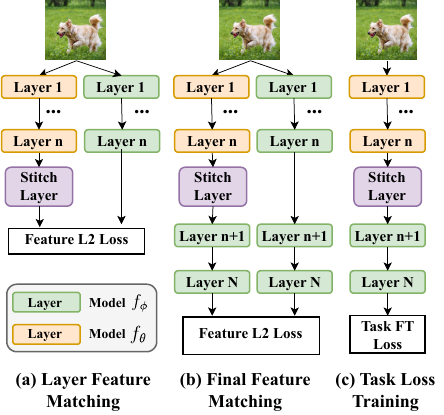}   
    \vspace{-1.8em}
    \caption{\small Model stitching training strategies: (a) Layer feature matching trains the stitch layer to match features between the source and target models at the stitch position. (b) Final feature matching trains the stitch layer so that the stitched model matches the target model’s final features. (c) Task loss training optimizes the stitch layer with downstream task objective. Across all strategies, the stitch layer is the only trainable component; the source and target models are kept frozen. }
    \label{fig:demo}
    \vspace{-0.5em}
\end{figure}

The last decade has seen a shift from crafting bespoke model architectures to pre-training vision foundation models (VFMs)---typically Transformers---on massive, heterogeneous data with diverse objectives~\cite{radford2021learning,caron2021emerging,zhai2023sigmoid}. VFMs now serve as default backbones for many tasks thanks to strong generalization and transferability: users can often fine-tune with little (or no) task-specific data and reach accuracy that was out of reach only a few years ago~\cite{mai2025lessons, mai2024fine}.

A natural question follows. Although VFMs achieve different scores on downstream, capability-probing tasks~\cite{tong2024eyes,fu2023mme}, are they fundamentally different ``end-to-end,'' or are their internal representations similar or compatible up to simple transformations? The answer matters for strategy: should we keep training new models (costly), or invest in managing and integrating existing ones?

We study this through model stitching: connecting the early layers of a source model to the later layers of a target model via a light stitch layer~\cite{bansal2021revisiting,csiszarik2021similarity} (\autoref{fig:demo}). Two models are stitchable if, with all original weights frozen and only the stitch layer trained, the stitched model matches the target model's accuracy (\ie, negligible drop). Existing results showed that small models (\eg, ResNet-18~\cite{he2016deep}) trained on the same dataset (\eg, CIFAR-10~\cite{krizhevsky2009learning}) are stitchable despite different initializations or objectives. We extend this question to VFMs, large Transformers trained with different datasets (\eg, LAION~\cite{ilharco2021openclip}, WebLI~\cite{zhai2023sigmoid}, LVD-142M~\cite{oquab2023dinov2}), objectives (contrastive vs.~reconstruction), and modalities (vision-language vs.~pure vision)---including CLIP~\cite{radford2021learning}, DINOv2~\cite{oquab2023dinov2}, DINOv3~\cite{simeoni2025dinov3} and SigLIP~2~\cite{zhai2023sigmoid}.

\noindent\textbf{Our contribution.} We develop a systematic protocol covering the stitch positions, stitch layer designs, training losses, and multiple downstream tasks, and conduct a comprehensive analysis.
Key insights are as follows.

\noindent\textbf{Naive stitching fails.} We revisit two common strategies to learn the stitch layer~\cite{bansal2021revisiting,smith2025functional}: 1) Layer Feature Matching---train the stitch layer to match features between the source and target models at the stitch position; and 2) Task-Loss Training---optimize only the downstream objective (\eg, cross-entropy). In the VFM setting, both struggle, sometimes even much worse than either constituent model, and failures are more pronounced at shallow stitch positions. 

\noindent\textbf{Tailored training matters.} A closer look at the failures suggests that how to train the stitch layer matters. On the one hand, low feature matching error at the stitch position does not imply aligned final representations, particularly for shallow stitches. On the other hand, Task-Loss Training faces an intrinsic optimization challenge at shallow stitch positions. As all target model's layers after the stitch positions are frozen, gradients originating from the prediction head with weak supervision (\eg, via the final pooled token) must traverse these frozen layers to adjust only the stitch layer, making the loss landscape poorly conditioned, especially when the stitch layer is randomly initialized. We therefore propose a simple two-stage recipe: (i) pre-train the stitch layer to match the target model's final output features (Final Feature Matching), and then (ii) fine-tune with the downstream task loss. The pre-training largely reduces output-feature discrepancy, and the subsequent fine-tuning turns the good initialization into a strong stitched model accuracy---often matching or surpassing linear probes of either VFM across stitch positions.

\noindent\textbf{Stitching integrates complementary strengths.} To ensure gains are not merely from the adding stitch layer capacity, we insert the same trainable module into the source-only and target-only models. The stitched model consistently outperforms these self-stitched controls across stitch positions (\autoref{fig:hybrid_vs_self}), indicating that stitching VFMs is not only feasible but also has the potential to fuse complementarity between VFMs. We verify this across datasets and tasks---classification and semantic segmentation.

\noindent\textbf{From probe to system: VFM Stitch Tree (VST).} Modern multimodal systems (\eg Multimodal LLMs and Vision-Language Models) increasingly deploy multiple VFMs to capture complementary visual cues but incurs linear compute/memory overhead. Leveraging stitchability, we propose the VFM Stitch Tree (VST): share early layers across VFMs and retain specialized deep layers, enabling a controllable accuracy-latency trade-off. Taking a LLaVA model~\cite{liu2023visual} with two VFMs as an example, VST with 22 shared layers and 1 specialized layer (4.3\% extra resources) achieves 45\% gain of the full two-VFM (requires 100\% extra resources). With 14 shared + 9 specialized layers (40\% extra), VST achieves 84\% of the gain. Thus, VST offers a compute-aware knob for integrating complementary VFMs in multimodal systems.

% Extending beyond pure vision tasks, we investigate multimodal systems. MLLMs often consume VFM tokens, and concatenating tokens from multiple VFMs can boost performance but incurs linear compute/memory overhead. Leveraging stitchability, we propose the VFM Stitch Tree (VST): share early layers across VFMs and retain specialized deep layers, enabling a controllable accuracy-latency trade-off. On a LLaVA model~\cite{xx} with two VFMs, VST with 22 shared layers and 1 specialized layer (4.3\% extra resources) achieves 45\% of the full two-VFM gain (which requires 100\% extra resources). With 14 shared + 9 specialized layers (40\% extra), VST achieves 84\% of the gain. Thus, VST offers a compute-aware knob for integrating complementary VFMs in multimodal systems.

\noindent\textbf{Summary.} Our paper has the following key contributions. 
\begin{itemize}
    \item We revisit model stitching for VFMs and show that with appropriate training, they are \emph{reliably stitchable}. Across vision tasks, stitched models consistently improve upon baseline performance, suggesting complementary knowledge transfer between VFMs.
    \item Based on the findings, we propose VFM Stitch Tree, offering a controllable performance–efficiency trade-off for multimodal LLMs that employ multiple VFMs.
    \item To our knowledge, this is the first systematic study of model stitching for VFMs, aiming to advance this technique from a pure diagnostic probe toward a practical recipe and open a path for future work to integrate complementary VFM strengths.
\end{itemize}

{

}

\section{Related Work}
\label{sec:related_work}

\subsection{Model Stitching and Representational Analysis}

\noindent \textbf{Foundations of Model Stitching.} Model stitching was introduced by \cite{lenc2015understanding} as a tool for analyzing representations through the lens of equivariance and equivalence. The key idea is to connect intermediate layers from two models using a trainable stitch layer and measure whether the stitched model remains functional. Subsequent work~\cite{bansal2021revisiting} showed that models trained under similar settings are often stitchable with negligible performance degradation, suggesting a certain degree of representational compatibility.  A critical refinement in this field is the distinction between representational similarity (\eg, CKA \cite{kornblith2019similarity}) and functional similarity (task performance compatibility)~\cite{csiszarik2021similarity}. More recently, \cite{smith2025functional} questioned whether stitching success should be interpreted as evidence of semantic alignment, arguing instead that it may largely reflect representational clustering.  \cite{klabunde2024similarity} provided a broader survey of representation comparison methods, highlighting the difficulty of drawing robust conclusions about neural representations.

% Model stitching was first introduced by \cite{lenc2015understanding} as a method for understanding image representations by measuring their equivariance and equivalence. The core idea involves connecting layers from different models through a trainable transformation to assess representational compatibility. Seminal work~\cite{bansal2021revisiting} extended this framework, demonstrating that models trained under similar conditions can be stitched together with minimal performance loss, demonstrating. \cite{csiszarik2021similarity} introduced the crucial distinction between representational similarity (statistical measures on embeddings like CKA~\cite{kornblith2019similarity} ) and functional similarity (task performance compatibility), demonstrating that high representational similarity does not necessarily predict successful task transfer, and conversely, functional compatibility can exist with relatively low CKA values. \cite{klabunde2024similarity} further surveyed these measures, highlighting the complexity of comparing neural network representations. Recent work by \cite{smith2025functional} challenged functional alignment interpretations, arguing that stitching success primarily reflects representational clustering rather than semantic similarity.

\noindent \textbf{Model Stitching Training.} Recent work has explored different objectives for learning the stitch layer. \cite{aaai2025tasklossmatch} compared layer feature matching, which trains the stitch layer to match source and target features at the stitch position, with task loss matching, which optimizes the stitch layer directly for downstream performance. They found that task loss matching can produce out-of-distribution intermediate representations that improve task-specific accuracy, whereas layer feature matching is often more reliable for representation analysis. Recently, Functional Latent Alignment~\cite{athanasiadis2025functional} was proposed to encourage alignment not only at the stitch position but also across subsequent layers.

% Model stitching has seen many variants in terms of how the stitch layer should be trained. \cite{aaai2025tasklossmatch} compared layer feature matching ( trains the stitch layer to match features between the
% source and target models at the stitch position) versus task loss matching (train the stitch layer with downstream task loss), demonstrating that task loss matching tends to create out-of-distribution representations to improve task-specific performance and layer feature matching is more robust for representation comparison.  \cite{athanasiadis2025functional} proposed Functional Latent Alignment to minimize not only the representations at the stitch position but also at the layers following it. 

\noindent \textbf{Application of Model Stitching.} Beyond representation analysis, model stitching has also been used for efficient model design and deployment. For example, \cite{pan2023stitchable} and its concurrent work ~\cite{yang2022deep} use stitching to construct flexible model families that support resource-adaptive inference by recombining components from different networks.

Despite this progress, existing work largely focuses on relatively small models trained from scratch or ImageNet-scale pretraining. In contrast, we explore model stitching between independently trained large VFMs with substantial heterogeneity in pretraining data (\eg, LAION~\cite{ilharco2021openclip}, LVD-142M~\cite{oquab2023dinov2}), learning objectives (contrastive versus reconstruction), and modality (vision-language versus pure vision).  Furthermore, we shift the paradigm from passive analysis to active knowledge fusion, demonstrating that stitching can bridge complementary specializations to surpass the performance of individual constituent models.

\subsection{Vision Foundation Models}
\label{sec:related_vfm}

A wide range of architectures, learning strategies and data have been explored to develop vision foundation models (VFMs). Self-supervised models such as DINOv2~\cite{oquab2023dinov2} learn strong visual representations without explicit labels, while vision-language models such as CLIP~\cite{radford2021learning} and SigLIP~\cite{zhai2023sigmoid, siglip2} use contrastive objectives to align images and text. Efforts to scale both model size and data \cite{sun2023eva, fang2023data} have further pushed the boundaries of zero-shot generalization and robust representation learning. As a result of these differences in architecture, supervision, and training data, different VFMs often exhibit distinct strengths and weaknesses across visual abilities, such as fine-grained recognition, depth estimation, and semantic grounding~\cite{mai2025ava, tong2024eyes}. This diversity motivates us to ask \textit{whether fundamentally different VFMs can be stitched together, and if so, how their complementary specializations can be leveraged efficiently.} Detailed descriptions of the VFMs considered in this work are provided in Appendix~\ref{sec_sup:exp_detail}.

\section{Model Stitching for VFMs}
\label{sec:method}
\subsection{Problem Formulation}

% (\orangecircle $\approx 0$) but high final feature distance (\orangerect $\approx 2.0-2.5$), while final matching maintains low final feature distance (\bluerect).
\begin{figure}[t]
\centering
\includegraphics[width=0.48\textwidth]{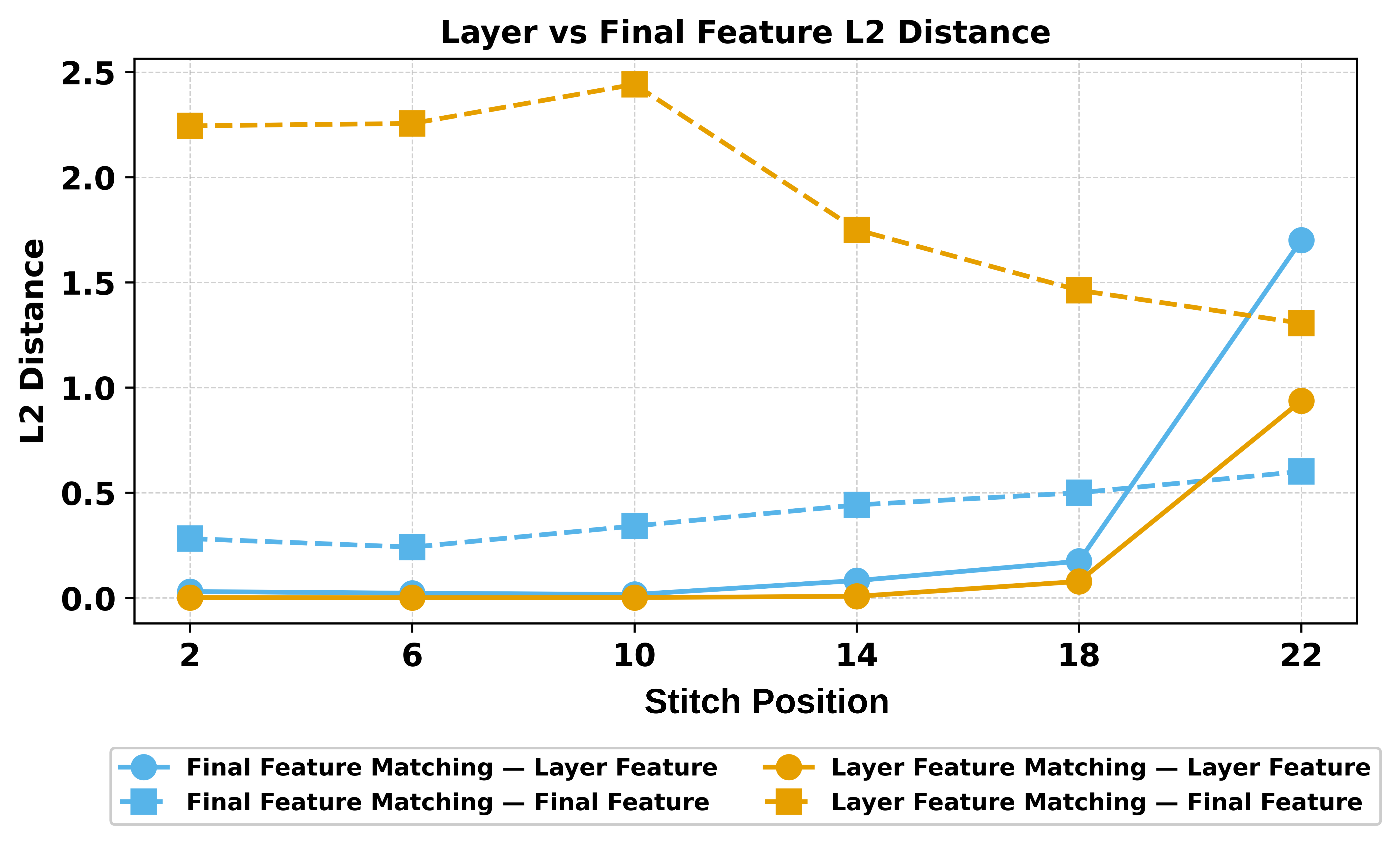}
\vspace{-2.em}
\caption{\small Feature distance of \textcolor{myorange}{Layer Feature Matching (LFM)} and \textcolor{myblue}{Final Feature Matching (FFM)} on SigLIP$\rightarrow$DINOv2. We measure the \(\ell_2\) distance between the stitched model’s features and those of the target model when stitching at different positions. \protect\greycircle indicates the \textit{layer} feature distance and  \protect\greyrect indicates the \textit{final} feature distance. While LFM minimizes the layer feature distance, it leaves a relatively large final feature distance. In contrast, FFM achieves substantially smaller final-feature distance. }
\label{fig:layer_vs_final}
\vspace{-0.5em}
\end{figure}

Let $f = f^N \circ \cdots \circ f^1$ be a VFM with $N$ Transformer layers.  For an image $x$, we define $R_n$ as the function to map the input $x$ to the intermediate features $\mA^n \in \R^{d\times k}$ at layer $n$: $R^n(x) = f^n \circ \cdots \circ f^1(x) = \mA^n$, where $d$ and $k$ denote the token dimension and number of token. Similarly, we define the function $T^N$ that map the $\mA^n$ to the output feature $\mA^N$: $T^N(\mA^n) = f^N \circ \cdots \circ f^n(\mA^n) = \mA^N$. 

Given a source and target VFM with parameters $\theta$ and  $\phi$ respectively, we can stitch them at layer $n$ by matching the source features $A^n_\theta$ to the target feature $A^n_\phi$ through a trainable stitching layer $S: \mathbb{R}^{{d_\theta}\times k} \rightarrow \mathbb{R}^{{d_\phi} \times k}$. Thus, the stitched VFM can be represented by:

\[
F(x) = T^N_{\phi} \circ S \circ R^n_{\theta}(x)
\]

We want to emphasize that all the parameters in $T^N_{\phi}$ and $ R^n_{\theta}$ are \textbf{frozen}; only the stitch layer $S$ is trainable. Following~\cite{csiszarik2021similarity, bansal2021revisiting}, for a given task, the source and target models are stitchable if the stitched model $F$ has limited performance degradation compared to the target model $f_\phi$. 

\subsection{How to Train the Stitch Layer?}
To answer whether heterogeneous VFMs are stitchable, we first explore the fundamental question: \textit{Are existing stitching approaches still effective for VFMs?} Two training strategies are most common (see \autoref{fig:demo}): (1) Layer Feature Matching~\cite{csiszarik2021similarity} and (2) Task Loss Training~\cite{bansal2021revisiting}. In the following sections, we aim to conduct controlled experiments to evaluate their compatibility with VFMs. 

\begin{figure}
    \centering
    \includegraphics[width=1\linewidth]{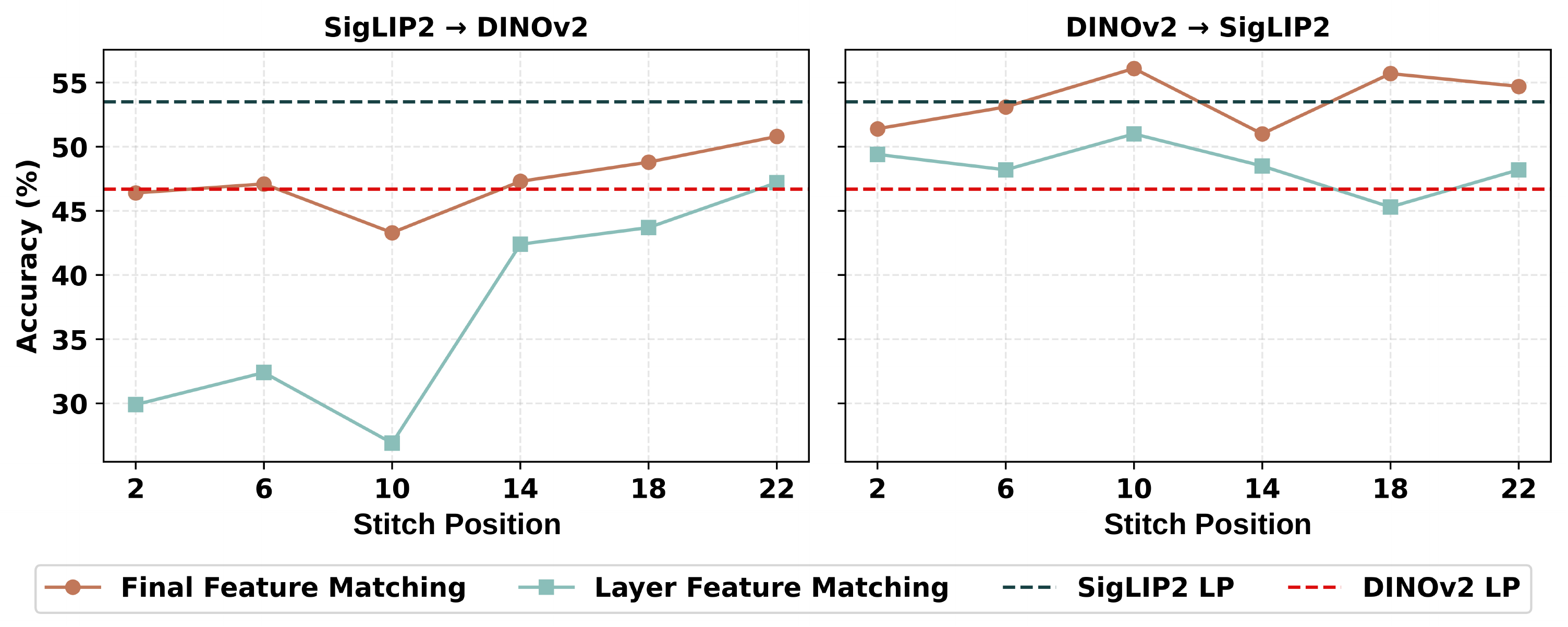}
    \vspace{-6mm}
    \caption{\small Final Feature Matching consistently shows better accuracy than Layer Feature Matching. In the DINOv2\(\rightarrow\)SigLIP2 case, the stitched model can even exceed the performance of both constituent models. }
    \label{fig:LP}
    \vspace{-0.5em}
\end{figure}

\mypar{Experiment Setup} We start with two commonly used VFMs: DINOv2-L~\cite{oquab2023dinov2} trained with only vision data (LVD-142M) in a self-supervised manner, and SigLIP2-L~\cite{siglip2} trained on vision-language data (WebLI)~\cite{chen2022pali} with both vision self-supervised loss and language supervised loss. Both VFMs have 24 layers. In the following control studies, we use a two-layer perceptron with ReLU (MLP) as the stitch layer, as it's a common feature projector for self-supervised learning~\cite{chen2020simple} and multimodal LLMs (\eg LLaVA-1.5~\cite{liu2024improved}). We focus first on image classification with fMoW~\cite{christie2018functional}, a challenging and commonly used fine-tuning dataset for VFM evaluation~\cite{fini2025multimodal}.  More setup details are provided in Appendix. Detailed investigation of stitch layer families,  datasets, VFM, and evaluation tasks will be discussed in the latter sections. 

\subsubsection{Layer Feature Matching (LFM)}
The intuition behind LFM is that the target features should be easily matched from the source features if they are similar enough. Given unlabeled training images $\mathcal{D}=\{x_i\}_{i=1}^{M}$, LFM trains $S$ by minimizing the feature discrepancy at layer $n$:
\begin{equation*}
\mathcal{L}_{LFM} = \frac{1}{M}\sum_{i=1}^M \|S(R^n_{\theta}(x_i)) - R^n_{\phi}(x_i)\|_2^2
\end{equation*}

\noindent Note that LFM training is label-free, and if we pre-extract the source and target features, the training can be done without VFM inference.

\mypar{Observations} To assess how well the target features can be transformed from source features across stitch positions $n \in [2, 6, 10, 14, 18, 22]$, we measure on the validation set the mean \(\ell_2\) \emph{layer} feature distance \(\|S(R^n_{\theta}(x)) - R^n_{\phi}(x)\|_2\) (\orangecircle\ in \autoref{fig:layer_vs_final}). As expected, because LFM optimizes this directly, these distances are very small (order of \(10^{-3}\)). However, small layer feature discrepancies do \emph{not} guarantee similarity at the final features. On the contrary, we observe substantially larger \emph{final} feature distances \(\|T^{N}_{\phi}(S(R^n_{\theta}(x))) - R^N_{\phi}(x)\|_2\) (\orangerect\ in \autoref{fig:layer_vs_final}), especially for shallow stitches. Intuitively, a small mismatch could accumulate and possibly be amplified by the frozen target layers, resulting in a pronounced final feature difference.

\mypar{Remedy} Motivated by the observations above and inspired by feature distillation, we propose \textbf{Final Feature Matching} (FFM), which trains the stitch layer to directly match the final features at layer $N$ between the stitch model $F$ and target model $f_\phi$: 

\begin{equation*}
\mathcal{L}_{FFM} = \frac{1}{M}\sum_{i=1}^M  \|T^N_{\phi}(S(R^n_{\theta}(x_i)))  - T^N_{\phi}(R^n_{\phi}(x_i))\|_2^2
\end{equation*}

\begin{figure}[]
\centering
\includegraphics[width=0.7\linewidth]{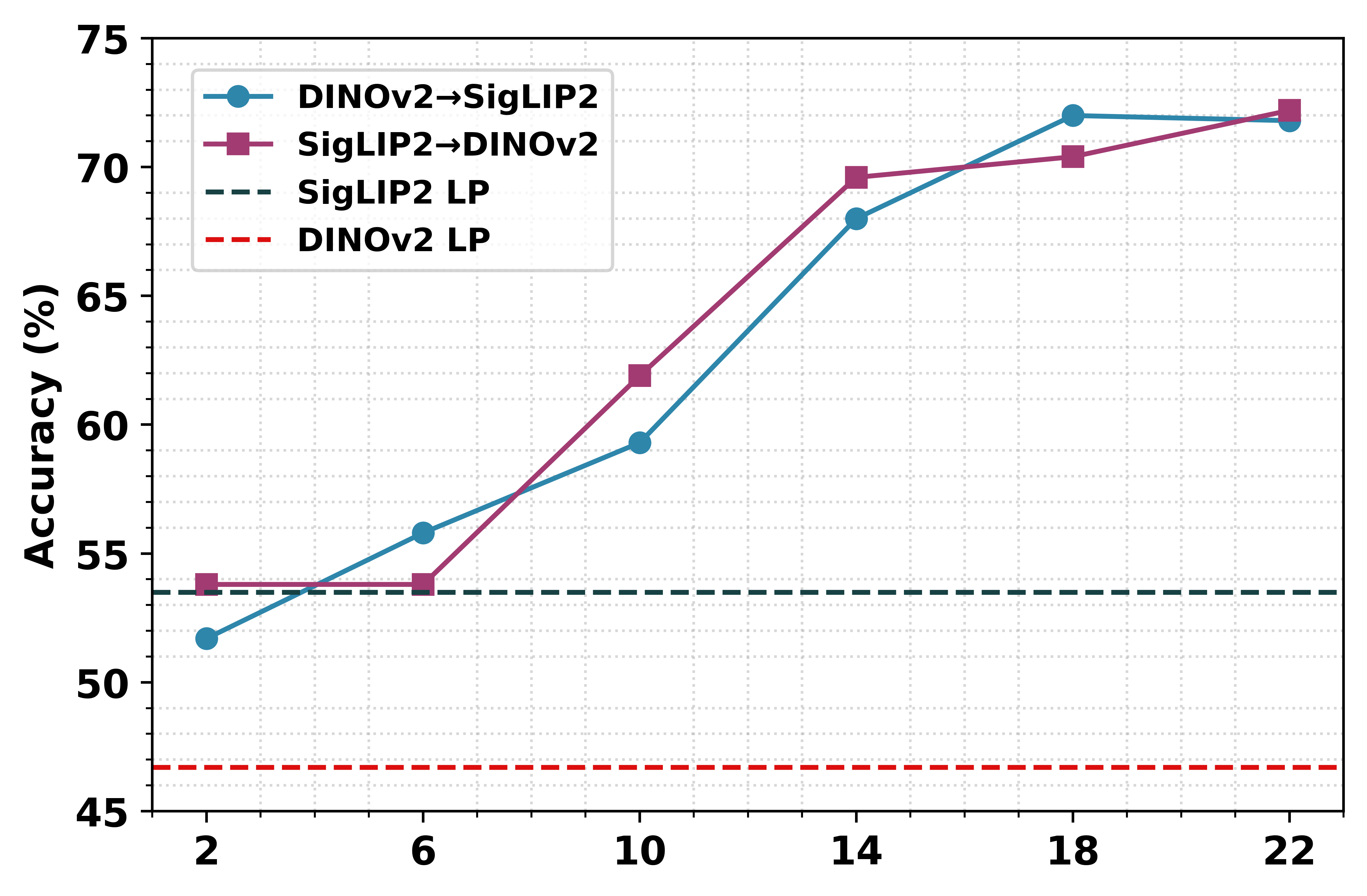}
\vspace{-2mm}
\caption{\small Our two-stage training approach (Final Feature Matching + Task Loss Training) allows stitched models to consistently outperform linear-probing of both constituent models. }
\label{fig:hybrid_lp}
\vspace{-0.5em}
\end{figure}

As shown in \autoref{fig:layer_vs_final}, compared with LFM, FFM significantly reduces the \emph{final} feature distances (\bluerect) at shallow stitch positions---precisely where difference accumulation is most pronounced. More interestingly, although FFM directly matches the final features, it still retains similarly low layer feature distances (\bluecircle) as LFM, suggesting that the supervision at the final layer can induce implicit local alignment at the stitch positions. The better final feature alignment translates into better linear probing accuracy (\autoref{fig:LP}): stitch models trained with FFM consistently outperform their LFM counterparts across stitch positions. Notably, in the DINOv2\(\rightarrow\)SigLIP2 case, the stitched model can exceed the performance of both constituent models under linear probing. Crucially, this improvement is achieved without using any labels; the stitch layer is trained only to match the final representations in FFM.

% To address this, we propose to match the output feature between the stitched and target model, which significantly reduces the output feature difference and shows much better linear probing accuracy (\autoref{fig:LP}). 

\subsubsection{Task Loss Training (TLT)}

Task Loss Training (TLT) trains the stitching layer directly with downstream task loss using the labeled training data $\mathcal{D} = \{(x_i, y_i)\}_{i=1}^M$, where $\ell$ is the task-specific loss (e.g., cross-entropy for classification):
\begin{equation}
\mathcal{L}_{task} = \frac{1}{M}\sum_{i=1}^M \ell(F(x_i), y_i)
\end{equation}

\mypar{Observations} Existing studies show that generally TLT outperforms LFM across stitch positions, as it directly optimizes the task objective. We also observe this pattern at deep stitch positions. However, in contrast to existing observations, we find that TLT fails sharply for \textbf{shallow} stitches. Taking DINOv2$\rightarrow$SigLIP2 stitching at layer 2 as an example, TLT attains only 25.1\%, markedly below the linear probing for DINOv2 (46.7\%),  SigLIP2 (53.5\%), LFM (49.4\%) and FFM (51.4\%).

\mypar{Remedy} 
When stitching shallowly, all post-stitch target layers are frozen; thus, gradients from the prediction head must backpropagate through a long, fixed transformation to update \emph{only} the stitch layer. This may attenuate update directions and cause optimization challenges, especially under random initialization and weak supervision (via pooled-token). To place the stitch layer in a better loss landscape, we adopt a simple two-stage procedure:
\begin{enumerate}[label=(\roman*)]
\item \textbf{Initialize} the stitch layer by pre-training with FFM
\item \textbf{Fine-tune} the stitch layer with TLT 
\end{enumerate}
As summarized in \autoref{tab:training_comparison}, FFM initialization not only rescues naïve TLT at shallow stitches where TLT alone underperforms, but also yields consistent gains at deeper stitch positions. In our experiments, it also significantly surpasses the linear-probing performance of both constituent models, as shown in \autoref{fig:hybrid_lp}. We also observe that shallow layers consistently underperform. We hypothesize that early layers may encode pretraining-specific features (e.g., low-level visual features for DINOv2, text-aligned features for SigLIP2) while deep layers may develop more transferable, task-relevant representations, leading to easier stitching than shallow layers. 

% \zd{stitch point analsis, from ke, Layer-dependent stitching performance using optimal configuration (MLP + Hybrid Final) on fMoW (accuracy in \%). Shallow layers (2-10) show poor performance, while deep layers (18-22) achieve strong results.
% \textbf{Finding: Deep layers enable effective stitching.} Shallow layers (2-10) consistently underperform, while deep layers (18-22) achieve optimal results. This suggests that:
% Early layers encode paradigm-specific features (e.g., edge detectors for DINOv2, text-aligned features for SigLIP), Deep layers develop more transferable, task-relevant representations, Optimal stitching occurs at layers 18-22 where representations are sufficiently abstract}

% With all post-stitch target layers frozen, gradients from the prediction head must back-propagate through a fixed and long (shallow stitch) transformation to update \textbf{only} the stitch layer. That means directions needed for improving the stitch layer can be severely attenuated, posing an optimization challenge under random initialization and via pooled-token supervision. To initialize the stitch layer to a better loss landscape,  propose a simple two-stage recipe: (i) pre-train the stitch layer with FFM; (ii) fine-tune it with the downstream task loss. As shown in \autoref{tab:training_comparison}, the FFM pre-training improves the naive TLT across stitch points for both stitch cases, and significantly outperforms linear probing of the source and target model. 

Our approach also resonates with a recent work~\cite{aaai2025tasklossmatch} which argues that TLT can create out-of-distribution representations that optimize for task performance at the cost of representational fidelity. Our FFM initialization mitigates this risk by first preserving representational fidelity, then applying task-level adaptation for functionality.

\begin{table}[t]
\centering
\small
\begin{tabular}{c|cccccc} \toprule Pre-train & \multicolumn{6}{c}{{DINOv2 $\rightarrow$ SigLIP2}} \\ \midrule Stitch Position & 2 & 6 & 10 & 14 & 18 & 22 \\ \midrule No & 25.1 & 39.4 & 52.6 & 62.3 & 68.6 & 68.6 \\ 
LFM & 33.1          & {57.8}          & {59.3}          & 67.3 & 68.4 & 70.3 \\
FFM & \textbf{51.7} & \textbf{59.8}  & \textbf{61.1} & \textbf{68.0} & \textbf{72.0} & \textbf{71.8} \\

\midrule Pre-train & \multicolumn{6}{c}{SigLIP2 $\rightarrow$ DINOv2} \\ \midrule Stitch Position & 2 & 6 & 10 & 14 & 18 & 22 \\ \midrule No & 38.7 & {50.7} & 58.3 & 64.4 & 70.4 & 70.1 \\ 
LFM & 35.8 & 51.8 & 59.2 & 69.4 & 70.2 & 72.1 \\ 
FFM & \textbf{53.8} & \textbf{53.8} & \textbf{61.9} & \textbf{69.6} & \textbf{70.4} & \textbf{72.2} \\ \bottomrule \end{tabular}

\vspace{-1mm}
\caption{\small Initializing the stitch layer with FFM substantially improves over naive TLT, especially at shallow stitch positions where naive TLT severely underperforms. FFM also provides consistent gains at deeper stitch positions. Compared with LFM initialization, FFM offers a clear advantage at shallow stitch positions, while their performance becomes similar at deeper stitch positions (Numbers are accuracy on fMoW). }
\label{tab:training_comparison}
\vspace{-0.5em}
\end{table}

\section{Where do Improvements Come From?}
\label{sec:self-stitch}

\noindent\textbf{Why do our results differ from prior works?}
With our two-stage training, the stitched model can substantially exceed linear probing for both the source and target models(\autoref{fig:hybrid_lp}). One might be tempted to conclude that VFMs are broadly stitchable, given that stitched models typically have around 0-10\% accuracy drop compared to the target model in prior works~\cite{bansal2021revisiting}. However, in earlier studies, the source, target, and stitch layer were all trained/evaluated on the \emph{same} dataset (e.g., CIFAR-10).  In contrast, VFMs are pretrained on massive, diverse data and evaluated via fine-tuning on downstream data. Under this regime, improvements could arise simply from \emph{task adaptation in the stitch layer}. We therefore design baselines that disentangle stitch layer capacity.

\subsection{Self-Stitch Baseline}
To rigorously test whether gains only stem from added capacity,  we introduce a Self-Stitch baseline: inserting the identical stitch layer into the source-only and target-only models at the same stitch positions. For example, for SigLIP2\(\rightarrow\)DINOv2 stitched at layer 22, the baselines are SigLIP2\(\rightarrow\)SigLIP2 and DINOv2\(\rightarrow\)DINOv2 with an identical stitch module at layer 22. Self-stitch controls share the \emph{same} trainable module, stitch positions, training loss, and downstream data. Matching self-stitch performance therefore indicates \emph{genuine VFM stitchability}.

\subsection{Stitching integrates complementary strengths}
\begin{figure}[t]
\centering
\includegraphics[width=0.8\linewidth]{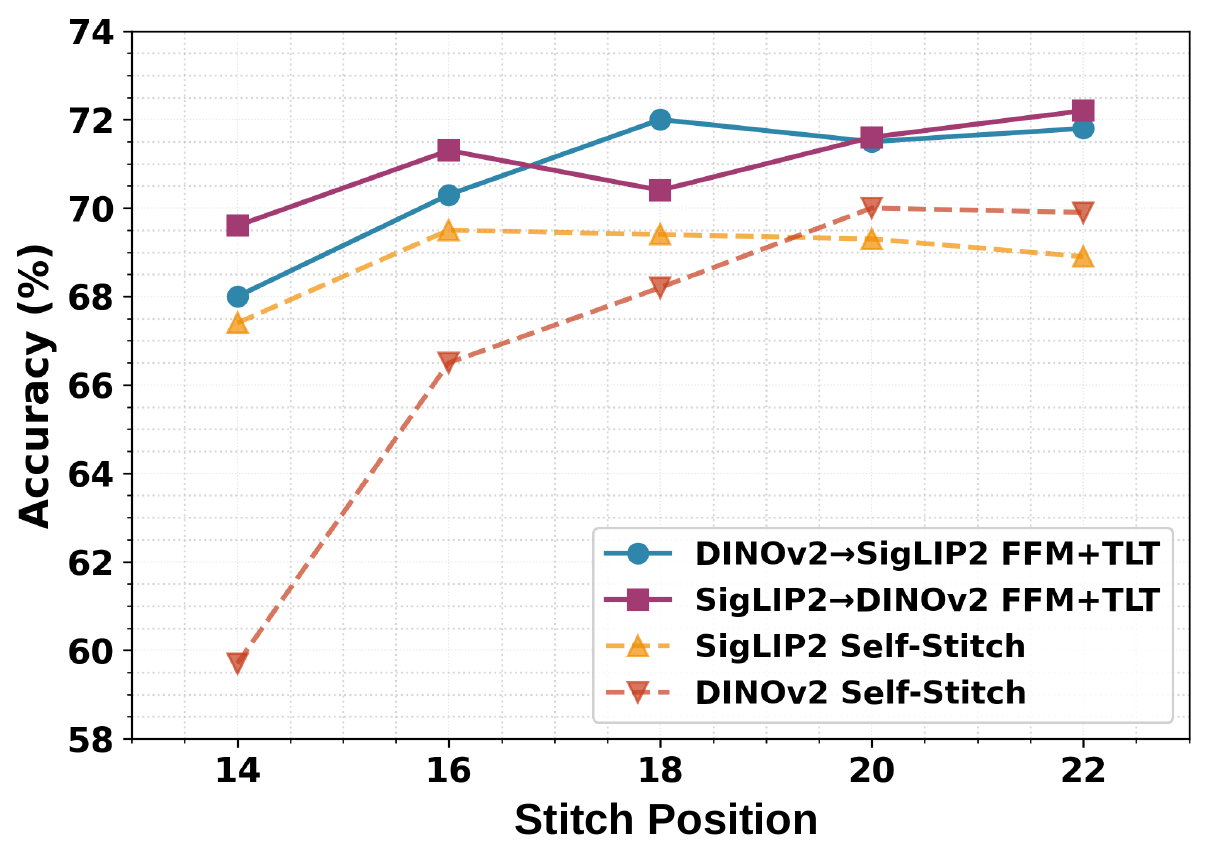}
\vspace{-0.7em}
\caption{\small Stitched Model vs Self-Stitch. Both DINOv2→SigLIP and SigLIP→DINOv2 (solid lines) consistently outperform their respective self-stitch baselines (dashed lines), demonstrating genuine knowledge fusion gains.}
\label{fig:hybrid_vs_self}
\vspace{-0.5em}
\end{figure}

\autoref{fig:hybrid_vs_self} compares the cross-VFM stitched model against their self-stitched baselines. Surprisingly, the stitched model consistently outperforms the self-stitched controls across stitch positions, indicating that improvements are \emph{not} explained solely by stitch layer capacity or downstream fine-tuning. Instead, the results suggest that stitching heterogeneous VFMs is not only feasible but also has the potential to \textit{fuse complementarity} between them.  To better understand how knowledge fusion affects prediction behavior, we provide prediction analysis for the stitched model and self-stitch baseline in Appendix~\ref{sec_sup:detail_results}.

\section{Generality of Model Stitching}
\label{sec:generalization}

To stay focused on exploring \textit{how to train the stitch layer} in previous sections, we adopt a fixed experiment setup. We assess the \emph{consistency} of our findings across datasets, tasks, stitch-layer architectures, and VFMs.

\subsection{Validation Across Datasets \& Tasks}

In addition to fMoW~\cite{christie2018functional} (satellite images), we evaluate on two widely used fine-grained classification datasets: iNat-Subset~\cite{van2018inaturalist} (animal species) and FGVC-Aircraft~\cite{maji2013fgvc}. Beyond classification, we evaluate semantic segmentation on ADE20K~\cite{zhou2017scene,zhou2019semantic} with a linear decoder~\cite{kerssies2024benchmark}. To simplify feature matching and avoid confounding factors, no data augmentation is used during training. 

 As shown in \autoref{tab:comprehensive_results}, stitched models consistently surpass the corresponding self-stitch baselines across datasets in classification (+0.7\% to +5.5\%), indicating that the observed knowledge fusion gain is \emph{not} a dataset-specific artifact but a generalizable observation.  In ADE20K, we likewise observe steady improvements over self-stitching (+0.5 to +0.7 mIoU), demonstrating that complementary knowledge fusion extends to dense prediction. We hypothesize that DINOv2 contributes robust perceptual structure while SigLIP2 provides stronger semantic alignment; their combination, mediated by the stitch layer, yields modest yet reliable improvements. More dataset details, experiment setup and results are provided in the Appendix.

\begin{table*}
\centering

\small
\begin{tabular}{lcccc}
\toprule
\multirow{2}{*}{} & \multicolumn{3}{c}{\textbf{Classification}} & \multicolumn{1}{c}{\textbf{Segmentation}} \\
\cmidrule(lr){2-4} \cmidrule(lr){5-5}
& \textbf{fMoW} & \textbf{iNaturalist} & \textbf{Aircraft} & \textbf{ADE20K} \\
\midrule
\textbf{Stitch Position} & \textbf{6} / \textbf{14} / \textbf{22} & \textbf{6} / \textbf{14} / \textbf{22} & \textbf{6} / \textbf{14} / \textbf{22} & \textbf{14} / \textbf{22} \\
\midrule
\textbf{DINOv2→DINOv2} & 41.5 / 59.7 / {69.9} & 56.9 / 81.5 / {91.2} & 37.8 / 79.3 / {91.2} & 35.4 / {50.9} \\
\textbf{SigLIP2→SigLIP2} & 50.5 / 62.0 / {68.9} & 71.2 / {88.5} / 87.3 & 67.9 / 88.1 / {89.3} & 44.5 / {50.5}  \\ 
\midrule
\textbf{DINOv2→SigLIP2} & \textbf{59.8} / 68.0 / {71.8} & 75.9 / 89.1 / \textbf{92.8} & 77.8 / 87.6 / \textbf{92.4} & 44.9 / {51.2} \\
\textbf{SigLIP2→DINOv2} & 53.8 / \textbf{69.6} / \textbf{72.2} & \textbf{86.3} / \textbf{88.9} / {91.9} & 80.7 / 89.0 / {91.0} & \textbf{49.0} / \textbf{51.4} \\
\bottomrule
\end{tabular}
\vspace{-2mm}
\caption{\small Comprehensive results across all datasets and tasks. Classification results in accuracy (\%), segmentation in mIoU (\%). All results use the two-stage approach (Final Feature Matching + Task Loss Training).}
\label{tab:comprehensive_results}
\vspace{-0.7em}
\end{table*}

\subsection{Validation Across More VFMs}
\label{sec: across_vfm}
\begin{figure}
    \centering
    \includegraphics[width=1.05\linewidth]{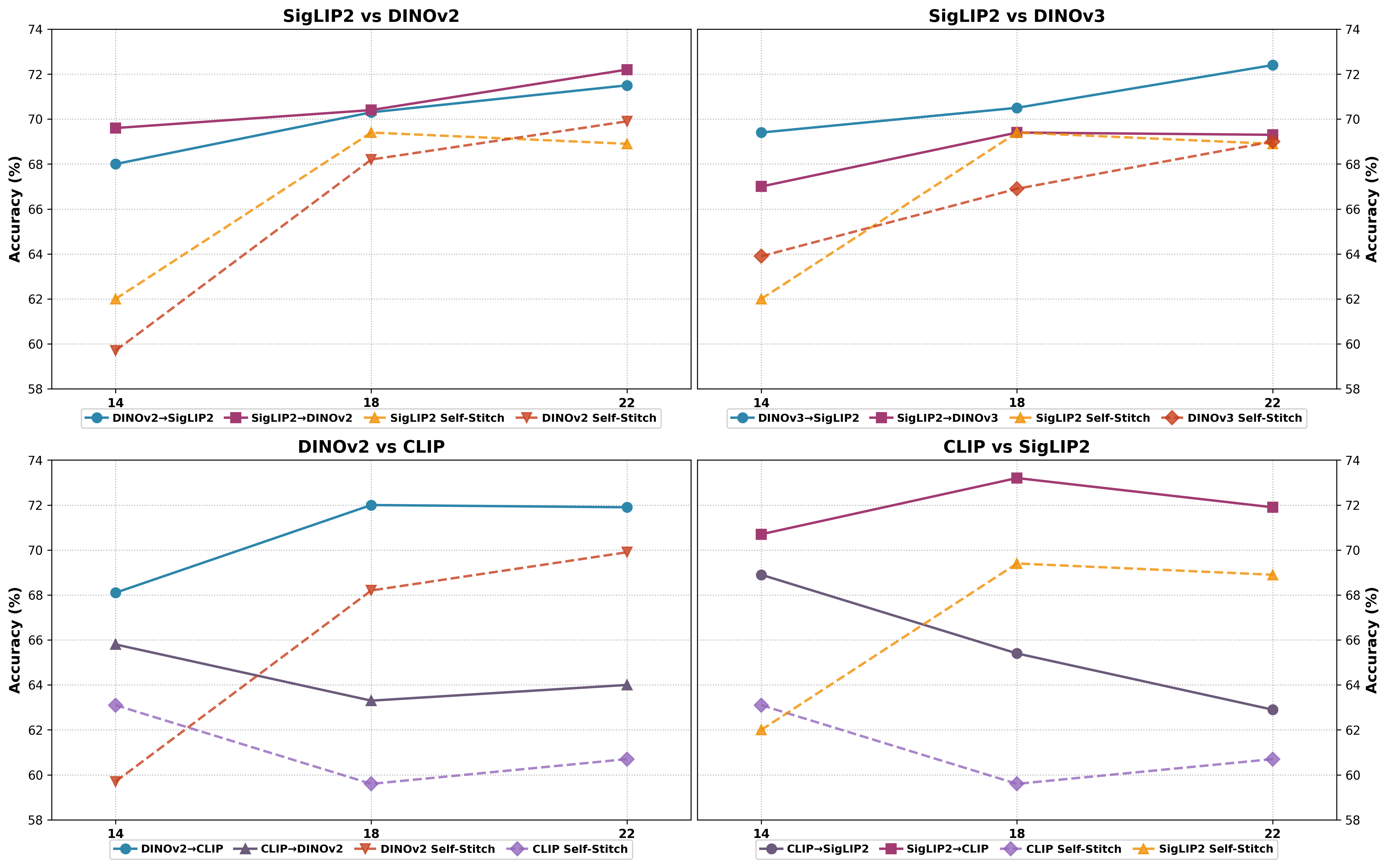}
    \caption{\small Evaluation of DINOv2, DINOv3, SigLIP2 and CLIP. The stitched models generally outperform both corresponding self-stitch baselines across stitched layers, except for using CLIP as the source model (discussion in \cref{sec: across_vfm}). }
    \label{fig:vfm_compare}
    \vspace{-0.7em}
\end{figure}

Beyond the DINOv2 and SigLIP2 pairs, we further include the widely used CLIP and the recently released DINOv3 to verify the generality of our findings. As shown in \autoref{fig:vfm_compare}, the stitched models generally outperform both corresponding self-stitch baselines across stitched layers, except for using CLIP as the source model. In this case, CLIP is a weaker VFM in our setting (as evidenced by lower linear-probing and self-stitch performance), so the stitched model can improve over CLIP itself but still fails to match the stronger target model. Viewing the source as an encoder, this suggests that a weak encoder may discard task-critical information that the stronger target network cannot recover. Conversely, when CLIP serves as the target, the stitched model achieves strong performance, indicating that as long as a strong source model preserves rich intermediate representations, even a relatively weak target can still transform these features into high-quality final predictions. These results are reminiscent of observations in encoder–decoder architectures for detection and segmentation, where upgrading only the encoder/backbone (e.g., from ResNet-50 to ResNet-101) reliably improves performance, while substituting a strong backbone with a weaker one consistently degrades accuracy even under an unchanged decoder~\cite{lin2017feature,kamann2020benchmarking,hou2024lmdeeplabv3plus}.

\begin{table}[H]
\centering
\small
\begin{tabular}{llcccccc}
\toprule
\multicolumn{2}{c}{Stitch Position} & \textbf{2} & \textbf{6} & \textbf{10} & \textbf{14} & \textbf{18} & \textbf{22} \\
\midrule
\multirow{3}{*}{D→S} 
& Linear & 26.1 & 54.3 & \textbf{59.5} & 66.5 & 69.1 & 69.6 \\
& MLP & \textbf{51.7} & \textbf{55.8} & 59.3 & \textbf{68.0} & \textbf{72.0} & \textbf{71.8} \\
& LoRA & 49.1 & 49.4 & 57.4 & 61.7 & 67.7 & 67.3 \\
\midrule
\multirow{3}{*}{S→D} 
& Linear & 50.3 & \textbf{56.4} & 60.0 & 65.7 & 69.6 & 71.9 \\
& MLP & \textbf{53.8} & 53.8 & 61.9 & \textbf{69.6} & \textbf{70.4} & \textbf{72.2} \\
& LoRA & 48.3 & 56.2 & \textbf{62.4} & 65.3 & 66.2 & 65.0 \\
\bottomrule
\end{tabular}
\vspace{-2mm}
\caption{\small Stitch layer comparison: MLP consistently outperforms Linear and the LoRA option across stitching directions (D→S, S→D) where D and S denote DINOv2 and SigLIP2, respectively. }
\label{tab:architecture_comparison}
\vspace{-0.3em}
\end{table}

\subsection{Across Stitch Layer Families}
Beyond the default MLP, we explore two different stitch layers: (1) a linear layer and (2) source model’s $n$ layer with LoRA~\cite{hu2021lora}, mapping \((n\!-\!1)\)-layer features of the source to the \(n\)-layer features of the target. Concretely, (2) can be represented as $F_{LoRA}(x) = T^N_{\phi} \circ f^n_{(\theta, LoRA)} \circ R^{n-1}_{\theta}(x)$ where only LoRA parameters within \(f^{n}_{\theta,\text{LoRA}}\)~\cite{hu2022lora, mai2025lessons} are trainable. From a capacity perspective, the linear layer processes tokens independently and is the least expressive; the MLP adds nonlinearity while still operating per token; the LoRA option allows inter-token interactions and is the most expressive. As reported in \autoref{tab:architecture_comparison}, the MLP generally attains the strongest performance on the benchmarks considered, with linear trailing—as expected given its lower capacity. Somewhat counterintuitively, the LoRA-based option often underperforms the MLP despite its higher expressiveness. A plausible interpretation is that stitching may benefit from \emph{controlled} mismatch that enables complementary information to be fused. If the stitch layer perfectly reproduces the target’s intermediate features, \(S(R^{n}_{\theta}(x)) = R^{n}_{\phi}(x)\), the stitched model would be expected to match the target’s self-stitch, leaving little room for cross-model complementarity.

\subsection{Does Stitching Require Task-Specific Data?}

Our previous results show that FFM is an effective stitching objective. However, these findings rely on task-specific training data (\eg use fMoW training images to learn the stitch layer for evaluation on fMoW). We therefore ask whether stitching must be task-specific, or whether it can instead be learned from general task-agnostic data. To study this, we train stitch layers using the task-agnostic LLaVA-1.5 data (1.1M images)~\cite{liu2024improved}, and evaluate the resulting stitched models on downstream tasks via linear probing, without further TLT.

\begin{table}[H]
\centering
\small
\setlength{\tabcolsep}{4pt}
\begin{tabular}{c|cc|cc}
\toprule
& \multicolumn{2}{c|}{fMoW} & \multicolumn{2}{c}{iNaturalist} \\
\cline{2-5}
& T-Spec. & T-Agn. & T-Spec. & T-Agn. \\
\midrule
D$\rightarrow$S & 54.7 & 53.8 & 77.3 & 77.45 \\
S$\rightarrow$D & 50.8 & 50.0 & 75.5 & 77.7 \\
\bottomrule
\end{tabular}
\vspace{-0.5em}
\caption{\small Comparison between task-specific (T-Spec.) and task-agnostic (T-Agn.) stitching. Results are reported for layer 22 (earlier analysis identifies as a strong stitch position).}
\label{tab:spe_agn}
\vspace{-0.5em}
\end{table}

As shown in \autoref{tab:spe_agn}, task-agnostic FFM achieves comparable, and in some cases better, accuracy compared to task-specific training. On fMoW, task-agnostic training matches task-specific performance, while on iNat, it surprisingly yields a performance gain. We hypothesize that iNat aligns more closely with the visual distribution of LLaVA-1.5 than fMoW, allowing the significantly larger scale of the LLaVA-1.5 data to facilitate more robust feature matching. These results suggest that stitch layers can be pre-trained in a task-agnostic manner and reused across diverse downstream applications, obviating the need for per-task retraining. More broadly, this indicates that stitching captures a fundamental alignment between the representation spaces of two VFMs rather than a narrow task-specific mapping. This opens the possibility for \textbf{Hybrid VFMs}: stitching can be used for synthesizing a better VFM by fusing the complementary strengths of distinct VFMs in a task-agnostic way. Such hybrid models can then serve as more versatile backbones for a wide array of downstream tasks.

\begin{tcolorbox}[colback=gray!5,colframe=gray!60,title= Key Findings]
\begin{itemize}
    \item Heterogeneous VFMs trained with different data, objectives, and modalities remain stitchable across diverse vision tasks and datasets.
    
    \item Stitched VFMs consistently outperform self-stitch baselines, indicating that the gains arise not merely from the added capacity of the stitch layer, but also from the complementary strengths of different VFMs.
\end{itemize}
\end{tcolorbox}

\section{From Insights To Applications}
\label{sec:application}

\begin{figure}[H]
        \centering
        \includegraphics[width=0.9\linewidth]{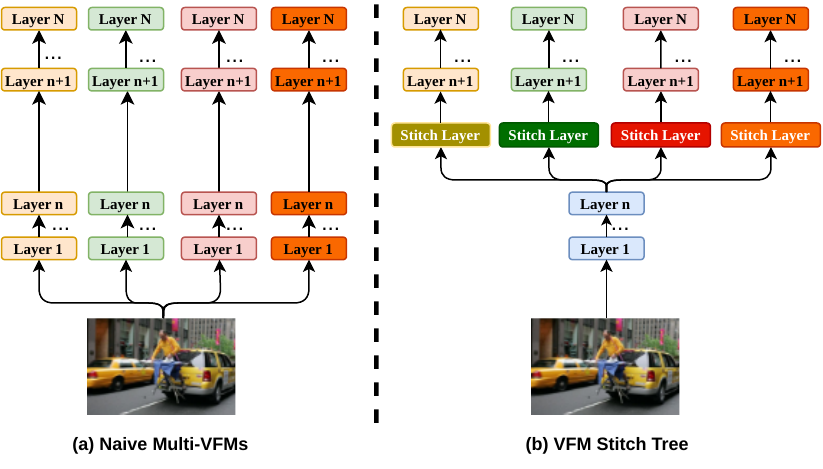}
        \vspace{-2mm}
        \caption{\small VFM Stitch Tree (VST) serves as a simple and efficient alternative to naively running all VFMs end-to-end.}
        \label{fig:vst_raw}
        \vspace{-0.5em}
\end{figure}

% Our findings on VFM stitchability open a fertile space of research questions about knowledge fusion while also indicating clear potential for practical utility. We illustrate a concrete example by improving efficiency in multi-VFM multimodal settings. 

\subsection{The Efficiency Challenge For Multi-VFMs}
Modern multimodal systems increasingly deploy multiple VFMs to capture complementary visual information. For example, MoF-LLaVA~\cite{tong2024eyes} combines CLIP and DINOv2 to preserve instruction following while improving visual grounding; OpenVLA~\cite{kim2024openvla} leverages SigLIP and DINOv2 to fuse low-level spatial structure with higher-level semantics; Cambrian-1~\cite{tong2024cambrian} shows that assembling four VFMs with distinct strengths can lift MLLM performance across diverse benchmarks. However, there is no free lunch: with \(k\) VFMs, one must load all of them into GPU memory ($k \times$) and process each input \(k\) times, yielding roughly \(k\times\) latency. \textbf{Question:} Can we design an architecture that retains the benefits of multiple VFMs \emph{without} incurring linear compute and memory costs?

\subsection{VFM Stitch Tree (VST)}

Motivated by the stitchability of VFMs, we propose the \emph{VFM Stitch Tree (VST)}: shares common shallow layers across VFMs while \emph{retains} model-specific deep layers, connected via stitch layers (\autoref{fig:vst_raw}). VST serves as a simple and efficient alternative to naively running all VFMs end-to-end. Consider Cambrian-1 with four VFMs. A VST stitched at layer 14 reduces GPU memory and computation relative to running all four full VFMs (\cref{tab:efficiency}).

\begin{table}[H]
\footnotesize
\centering

\begin{tabular}{l|cccc}
\toprule
\textbf{Method} & \textbf{Param (M)} & \textbf{GFLOPs} & \textbf{Lat. (ms)} & \textbf{Mem (MB)} \\
\midrule
Naive  & 1581.4 & 878.8 & 96.2 & 6075 \\
VST-6  & 1219.0 & 701.7 & 71.4 & 4502 \\
VST-14 & 815.9  & 419.5 & 43.9 & 3065 \\
VST-22 & 412.9  & 217.4 & 20.6 & 1527 \\
\bottomrule
\end{tabular}
\vspace{-0.8em}
\caption{Efficiency metrics for different stitching configurations. Latency is measured on a single NVIDIA A100 GPU with a batch size of 1. VST significantly reduces computational overhead compared to the Naive Multi-VFM baseline.}
\label{tab:efficiency}
\vspace{-0.8em}
\end{table}

\begin{figure}
    \centering
    \includegraphics[width=1\linewidth]{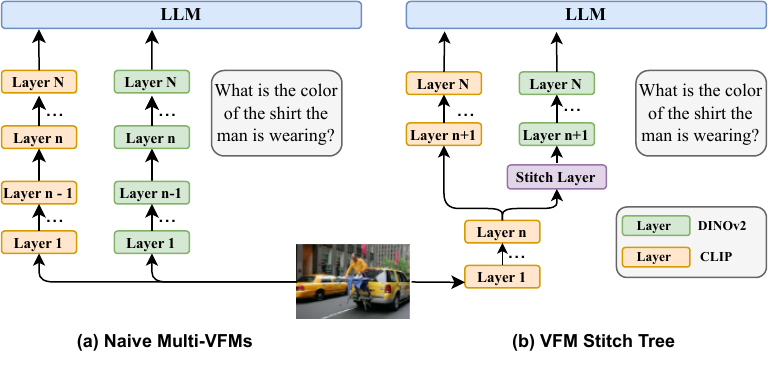}
    \vspace{-6mm}
    \caption{\small VFM Stitch Tree (VST) can be easily applied in various multimodal systems. This is an example of VST in a MoF-LLaVA with CLIP and DINOv2.   }
    \label{fig:stitch-tree}
    \vspace{-0.5em}
\end{figure}

\begin{figure}
    \centering
    \includegraphics[width=0.9\linewidth]{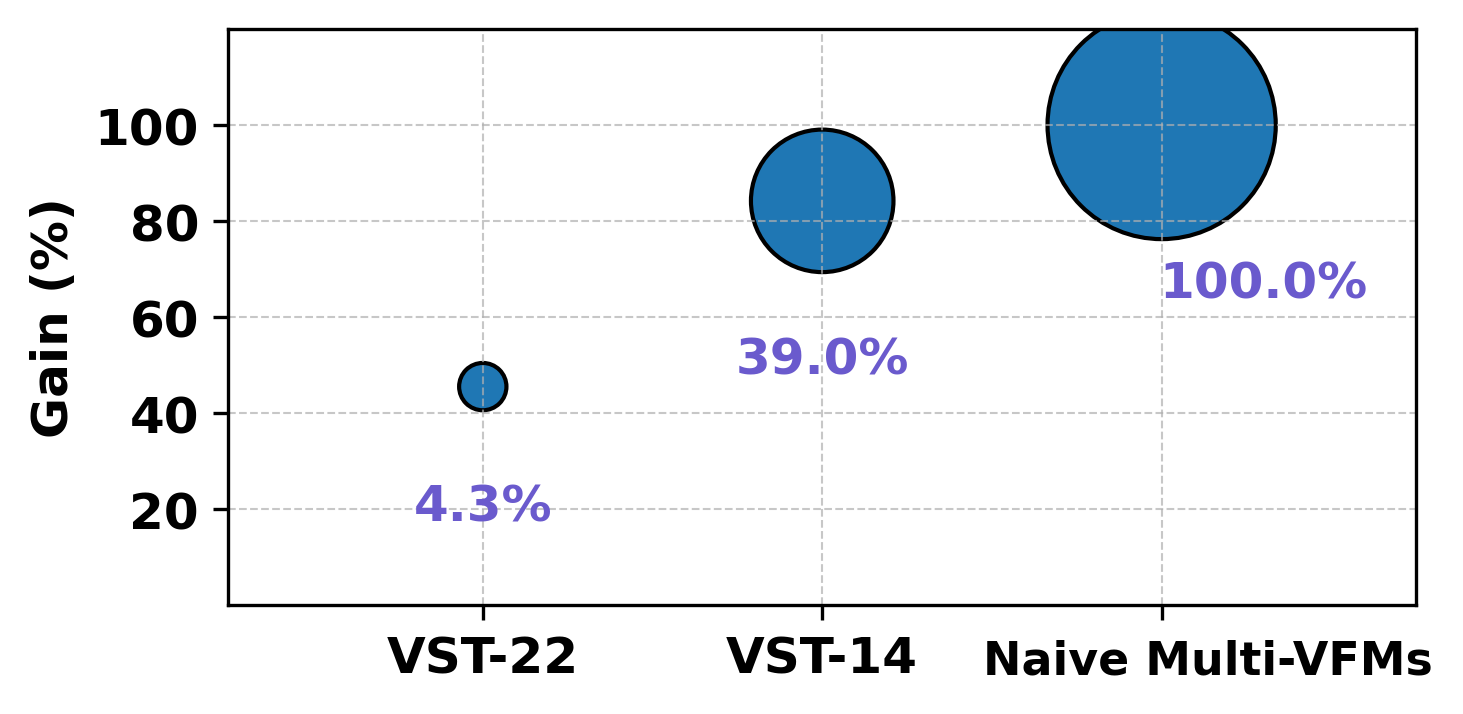}
    \vspace{-3mm}
    \caption{\small  VST recovers 45\% and 84\% of the gain achieved by the naive multi-VFM baseline (100\% extra resources), with only 4.3\% and 39\% extra resources. Details are in Appendix.  }
    \label{fig:resource}
    \vspace{-0.5em}
\end{figure}

We instantiate VST within a MoF-LLaVA (CLIP + DINOv2) with a Qwen-3B LLM~\cite{hui2024qwen2}. The model is trained following the standard LLaVA-1.5~\cite{liu2024improved} training recipe and datasets. We evaluate VST stitched at layer 14 and 22, as demonstrated in \autoref{fig:stitch-tree}. Relative to a single-VFM LLaVA baseline, the naïve two-VFM MoF-LLaVA doubles the VFM cost (\( 100\%\) extra). In contrast, our VST variants require only \( 39\%\) (VST-14) and \( 4.3\%\) (VST-22) additional resources. In this early exploration, we evaluate the MLLM on VQAv2~\cite{goyal2017making} and MME-Perception and MME-Cognition~\cite{fu2025mme}. As shown in \autoref{fig:resource}, VST-22, with just \emph{one} additional specialized layer (\( 4.3\%\) overhead), recovers \(45\%\) of the two-VFM performance gain; when more budget is available, VST-14 reaches \( 84\%\) of that gain at \(39\%\) overhead. We provide experiment setup, detailed results and resource calculation in the Appendix.

\mypar{Accuracy–efficiency knob} Without VST, users face a binary choice: deploy an entire additional VFM (higher performance, \(100\%\) extra backbone cost) or none (no gain). VST introduces a \emph{compute-aware knob} that interpolates between these extremes, enabling controllable performance–efficiency trade-offs under real deployment constraints. We view this as an initial step toward resource-conscious multi-VFM design; broader evaluations are provided in the Appendix.

We want to highlight that our findings on VFM stitchability open a fertile space of research questions about knowledge fusion and also indicate potential for practical utility. VST is just \emph{one} instantiation of how these insights can be applied. The central takeaway is that heterogeneous VFMs can be stitched and fused to exploit complementary knowledge, which has potential applications across different domains and use cases.

\section{Conclusion}
We revisit model stitching under the foundation model regime and find that, contrary to the common assumption that heterogeneity hinders compatibility, VFMs trained with different data, objectives, and modalities can in fact be stitched effectively. Our study shows that the key lies in how the stitch layer is trained: while conventional layer feature matching and naive end-to-end task optimization often fail, especially for shallow stitches, a simple two-stage strategy based on final feature matching followed by task fine-tuning yields strong and consistent results. Importantly, stitched models not only remain functional but often surpass corresponding self-stitch baselines, suggesting that stitching can actively fuse complementary knowledge across VFMs rather than merely preserve target behavior. We further demonstrate that these findings generalize across tasks, datasets, stitch-layer designs, and model families, and that stitch layers can even be pretrained in a task-agnostic manner. Finally, we show how these insights enable VFM Stitch Tree, a practical architecture that reduces the cost of multi-VFM systems while preserving much of their benefit. Taken together, our work advances model stitching from a representational probe to a practical paradigm for integrating and scaling heterogeneous foundation models.

% We revisited model stitching in the foundation model era. Our systematic study across diverse objectives, datasets, and modality mixes shows that the answer is yes, provided that we train the stitch layer appropriately. 

% We find that conventional strategies, such as matching intermediate activations at the stitch point or directly optimizing task loss, often fail to preserve accuracy, especially at shallow stitch points. In contrast, a simple feature-matching loss at the target model’s penultimate layer makes heterogeneous VFMs reliably stitchable across a range of vision tasks, with stitched models consistently improving over strong self-stitch baselines and revealing complementary knowledge transfer between VFMs.

% Building on these empirical insights, we introduced the VFM Stitch Tree (VST), an architecture that shares early layers while retaining the later layers of multiple VFMs, enabling a controllable accuracy–latency trade-off for multimodal LLMs that rely on more than one visual encoder.  Taken together, our results elevate stitching from a purely diagnostic tool to a practical recipe for integrating complementary VFM strengths and for pinpointing where their representations align or diverge. We hope this work provides both a blueprint and a testbed for future research on composing, reusing, and scaling VFMs within broader multimodal systems.

\newpage
\clearpage
{\small
\bibliographystyle{ieeenat_fullname}
\bibliography{main}

@inproceedings{bansal2021revisiting,
  title        = {Revisiting Model Stitching to Compare Neural Representations},
  author       = {Bansal, Yamini and Nakkiran, Preetum and Barak, Boaz},
  booktitle    = {Advances in Neural Information Processing Systems (NeurIPS)},
  year         = {2021}
}

@inproceedings{smith2025functional,
  title        = {Functional Alignment Can Mislead: Examining Model Stitching},
  author       = {Smith, Damian and Mannering, Harvey and Marcu, Antonia},
  booktitle    = {Proceedings of the 42nd International Conference on Machine Learning (ICML)},
  year         = {2025}
}

@article{mai2025ava,
  title={Ava-bench: Atomic visual ability benchmark for vision foundation models},
  author={Mai, Zheda and Chowdhury, Arpita and Wang, Zihe and Jeon, Sooyoung and Wang, Lemeng and Hou, Jiacheng and Chao, Wei-Lun},
  journal={arXiv preprint arXiv:2506.09082},
  year={2025}
}

@inproceedings{lenc2015understanding,
  title        = {Understanding Image Representations by Measuring Their Equivariance and Equivalence},
  author       = {Lenc, Karel and Vedaldi, Andrea},
  booktitle    = {Proceedings of the IEEE Conference on Computer Vision and Pattern Recognition (CVPR)},
  year         = {2015}
}

@article{hu2022lora,
  title={Lora: Low-rank adaptation of large language models.},
  author={Hu, Edward J and Shen, Yelong and Wallis, Phillip and Allen-Zhu, Zeyuan and Li, Yuanzhi and Wang, Shean and Wang, Liang and Chen, Weizhu and others},
  journal={Iclr},
  volume={1},
  number={2},
  pages={3},
  year={2022}
}

@inproceedings{mai2025lessons,
  title={Lessons and insights from a unifying study of parameter-efficient fine-tuning (peft) in visual recognition},
  author={Mai, Zheda and Zhang, Ping and Tu, Cheng-Hao and Chen, Hong-You and Nguyen, Quang-Huy and Zhang, Li and Chao, Wei-Lun},
  booktitle={Proceedings of the Computer Vision and Pattern Recognition Conference},
  pages={14845--14857},
  year={2025}
}

@article{mai2024fine,
  title={Fine-tuning is fine, if calibrated},
  author={Mai, Zheda and Chowdhury, Arpita and Zhang, Ping and Tu, Cheng-Hao and Chen, Hong-You and Pahuja, Vardaan and Berger-Wolf, Tanya and Gao, Song and Stewart, Charles and Su, Yu and others},
  journal={Advances in neural information processing systems},
  volume={37},
  pages={136084--136119},
  year={2024}
}

@article{simeoni2025dinov3,
  title        = {DINOv3},
  author       = {Sim{\'e}oni, Oriane and Vo, Huy V. and Seitzer, Maximilian and Baldassarre, Federico and Oquab, Maxime and Jose, Cijo and Khalidov, Vasil and Szafraniec, Marc and Yi, Seungeun and Ramamonjisoa, Micha{\"e}l and Massa, Francisco and Haziza, Daniel and Wehrstedt, Luca and Wang, Jianyuan and Darcet, Timoth{\'e}e and Moutakanni, Th{\'e}o and Sentana, Leonel and Roberts, Claire and Brandt, John and Couprie, Camille and Mairal, Julien and J{\'e}gou, Herv{\'e} and Labatut, Patrick and Bojanowski, Piotr},
  journal      = {arXiv preprint arXiv:2508.10104},
  year         = {2025}
}

@article{siglip2,
  title={Siglip 2: Multilingual vision-language encoders with improved semantic understanding, localization, and dense features},
  author={Tschannen, Michael and Gritsenko, Alexey and Wang, Xiao and Naeem, Muhammad Ferjad and Alabdulmohsin, Ibrahim and Parthasarathy, Nikhil and Evans, Talfan and Beyer, Lucas and Xia, Ye and Mustafa, Basil and others},
  journal={arXiv preprint arXiv:2502.14786},
  year={2025}
}

@inproceedings{radford2021learning,
  title        = {Learning Transferable Visual Models from Natural Language Supervision},
  author       = {Radford, Alec and Kim, Jong Wook and Hallacy, Chris and Ramesh, Aditya and Goh, Gabriel and Agarwal, Sandhini and Sastry, Girish and Askell, Amanda and Mishkin, Pamela and Clark, Jack and others},
  booktitle    = {Proceedings of the 38th International Conference on Machine Learning (ICML)},
  year         = {2021}
}

@article{oquab2023dinov2,
  title        = {DINOv2: Learning Robust Visual Features without Supervision},
  author       = {Oquab, Maxime and Darcet, Timoth{\'e}e and Moutakanni, Th{\'e}o and Vo, Huy V and Szafraniec, Marc and Khalidov, Vasil and Fernandez, Pierre and Haziza, Daniel and Massa, Francisco and El-Nouby, Alaaeldin and others},
  journal      = {arXiv preprint arXiv:2304.07193},
  year         = {2023}
}

@article{zhai2023sigmoid,
  title        = {Sigmoid Loss for Language-Image Pre-Training},
  author       = {Zhai, Xiaohua and Mustafa, Basil and Kolesnikov, Alexander and Beyer, Lucas},
  journal      = {arXiv preprint arXiv:2303.15343},
  year         = {2023}
}

@inproceedings{kornblith2019similarity,
  title={Similarity of neural network representations revisited},
  author={Kornblith, Simon and Norouzi, Mohammad and Lee, Honglak and Hinton, Geoffrey},
  booktitle={International conference on machine learning},
  pages={3519--3529},
  year={2019},
  organization={PMlR}
}

@article{yang2022deep,
  title={Deep model reassembly},
  author={Yang, Xingyi and Zhou, Daquan and Liu, Songhua and Ye, Jingwen and Wang, Xinchao},
  journal={Advances in neural information processing systems},
  volume={35},
  pages={25739--25753},
  year={2022}
}

@inproceedings{athanasiadis2025functional,
  title={Functional Similarity by Functional Latent Alignment},
  author={Athanasiadis, Ioannis and Karmush, Anmar and Felsberg, Michael},
  booktitle={Greeks in AI Symposium 2025},
  year={2025}
}

@inproceedings{liu2023visual,
  title        = {Visual Instruction Tuning},
  author       = {Liu, Haotian and Li, Chunyuan and Wu, Qingyang and Lee, Yong Jae},
  booktitle    = {Advances in Neural Information Processing Systems (NeurIPS)},
  year         = {2023}
}

@article{liu2024improved,
  title        = {Improved Baselines with Visual Instruction Tuning},
  author       = {Liu, Haotian and Li, Chunyuan and Li, Yong Jae and Lee, Yong Jae},
  journal      = {arXiv preprint arXiv:2310.03744},
  year         = {2024}
}

@inproceedings{bai2023qwen,
  title        = {Qwen Technical Report},
  author       = {Bai, Jinze and Bai, Shuai and Chu, Yunfei and Cui, Zeyu and Dang, Kai and Deng, Xiaodong and Fan, Yang and Ge, Wenbin and Han, Yu and Huang, Fei and others},
  booktitle    = {arXiv preprint arXiv:2309.16609},
  year         = {2023}
}

@inproceedings{fu2023mme,
  title        = {MME: A Comprehensive Evaluation Benchmark for Multimodal Large Language Models},
  author       = {Fu, Chaoyou and Chen, Peixian and Shen, Yunhang and Qin, Yulei and Zhang, Mengdan and Lin, Xu and Yang, Jinrui and Zheng, Xiawu and Li, Ke and Sun, Xing and others},
  booktitle    = {arXiv preprint arXiv:2306.13394},
  year         = {2023}
}

@article{christie2018functional,
  title        = {Functional Map of the World},
  author       = {Christie, Gordon and Fendley, Neil and Wilson, James and Mukherjee, Ryan},
  journal      = {Proceedings of the IEEE Conference on Computer Vision and Pattern Recognition (CVPR)},
  year         = {2018}
}

@article{tong2024cambrian,
  title={Cambrian-1: A fully open, vision-centric exploration of multimodal llms},
  author={Tong, Peter and Brown, Ellis and Wu, Penghao and Woo, Sanghyun and IYER, Adithya Jairam Vedagiri and Akula, Sai Charitha and Yang, Shusheng and Yang, Jihan and Middepogu, Manoj and Wang, Ziteng and others},
  journal={Advances in Neural Information Processing Systems},
  volume={37},
  pages={87310--87356},
  year={2024}
}

@article{kim2024openvla,
  title={Openvla: An open-source vision-language-action model},
  author={Kim, Moo Jin and Pertsch, Karl and Karamcheti, Siddharth and Xiao, Ted and Balakrishna, Ashwin and Nair, Suraj and Rafailov, Rafael and Foster, Ethan and Lam, Grace and Sanketi, Pannag and others},
  journal={arXiv preprint arXiv:2406.09246},
  year={2024}
}

@inproceedings{chen2020simple,
  title={A simple framework for contrastive learning of visual representations},
  author={Chen, Ting and Kornblith, Simon and Norouzi, Mohammad and Hinton, Geoffrey},
  booktitle={International conference on machine learning},
  pages={1597--1607},
  year={2020},
  organization={PmLR}
}

@inproceedings{zhou2019semantic,
  title        = {Semantic Understanding of Scenes through the ADE20K Dataset},
  author       = {Zhou, Bolei and Zhao, Hang and Puig, Xavier and Xiao, Tete and Fidler, Sanja and Barriuso, Adela and Torralba, Antonio},
  booktitle    = {International Journal of Computer Vision},
  year         = {2019}
}

@inproceedings{zhou2017scene,
  title        = {Scene Parsing through ADE20K Dataset},
  author       = {Zhou, Bolei and Zhao, Hang and Puig, Xavier and Fidler, Sanja and Barriuso, Adela and Torralba, Antonio},
  booktitle    = {Proceedings of the IEEE Conference on Computer Vision and Pattern Vision (CVPR)},
  year         = {2017}
}

@inproceedings{csiszarik2021similarity,
  title        = {Similarity and Matching of Neural Network Representations},
  author       = {Csisz\'arik, Adri\'an and K\H{o}r\"osi-Szab\'o, P\'eter and Matszangosz, \'Akos K. and Papp, Gergely and Varga, D\'aniel},
  booktitle    = {Advances in Neural Information Processing Systems (NeurIPS)},
  volume       = {34},
  pages        = {5656--5668},
  year         = {2021}
}

@inproceedings{klabunde2024similarity,
  title        = {Similarity of Neural Network Models Revisited: Measuring Functional Similarity},
  author       = {Klabunde, Max and Schubert, Tobias and Lapuschkin, Sebastian},
  booktitle    = {International Conference on Machine Learning (ICML)},
  year         = {2024}
}

@article{hui2024qwen2,
  title={Qwen2. 5-coder technical report},
  author={Hui, Binyuan and Yang, Jian and Cui, Zeyu and Yang, Jiaxi and Liu, Dayiheng and Zhang, Lei and Liu, Tianyu and Zhang, Jiajun and Yu, Bowen and Lu, Keming and others},
  journal={arXiv preprint arXiv:2409.12186},
  year={2024}
}

@article{tong2024eyes,
  title        = {Eyes Wide Shut? Exploring the Visual Shortcomings of Multimodal LLMs},
  author       = {Tong, Shengbang and Liu, Zhuang and Zhai, Yuexiang and Ma, Yi and LeCun, Yann and Xie, Saining},
  journal      = {arXiv preprint arXiv:2401.06209},
  year         = {2024}
}

@inproceedings{he2016deep,
  title        = {Deep Residual Learning for Image Recognition},
  author       = {He, Kaiming and Zhang, Xiangyu and Ren, Shaoqing and Sun, Jian},
  booktitle    = {Proceedings of the IEEE Conference on Computer Vision and Pattern Recognition (CVPR)},
  pages        = {770--778},
  year         = {2016}
}

@inproceedings{fu2025mme,
  title={MME: A comprehensive evaluation benchmark for multimodal large language models},
  author={Fu, Chaoyou and Chen, Peixian and Shen, Yunhang and Qin, Yulei and Zhang, Mengdan and Lin, Xu and Yang, Jinrui and Zheng, Xiawu and Li, Ke and Sun, Xing and others},
  booktitle={The Thirty-ninth Annual Conference on Neural Information Processing Systems Datasets and Benchmarks Track},
  year={2025}
}

@inproceedings{caron2021emerging,
  title        = {Emerging Properties in Self-Supervised Vision Transformers},
  author       = {Caron, Mathilde and Touvron, Hugo and Misra, Ishan and J{\'e}gou, Herv{\'e} and Mairal, Julien and Bojanowski, Piotr and Joulin, Armand},
  booktitle    = {Proceedings of the IEEE/CVF International Conference on Computer Vision (ICCV)},
  pages        = {9650--9660},
  year         = {2021}
}

@misc{ilharco2021openclip,
  title        = {OpenCLIP},
  author       = {Ilharco, Gabriel and Wortsman, Mitchell and Wightman, Ross and Gordon, Cade and Carlini, Nicholas and Taori, Rohan and Dave, Achal and Shankar, Vaishaal and Namkoong, Hongseok and Miller, John and others},
  howpublished = {\url{https://github.com/mlfoundations/open_clip}},
  year         = {2021}
}

@inproceedings{van2018inaturalist,
  title        = {The iNaturalist Species Classification and Detection Dataset},
  author       = {Van Horn, Grant and Mac Aodha, Oisin and Song, Yang and Cui, Yin and Sun, Chen and Shepard, Alex and Adam, Hartwig and Perona, Pietro and Belongie, Serge},
  booktitle    = {Proceedings of the IEEE Conference on Computer Vision and Pattern Recognition (CVPR)},
  pages        = {8769--8778},
  year         = {2018}
}

@misc{maji2013fgvc,
  title        = {Fine-Grained Visual Classification of Aircraft},
  author       = {Maji, Subhransu and Rahtu, Esa and Kannala, Juho and Blaschko, Matthew and Vedaldi, Andrea},
  howpublished = {\url{http://www.robots.ox.ac.uk/~vgg/data/fgvc-aircraft/}},
  note         = {arXiv preprint arXiv:1306.5151},
  year         = {2013}
}

@inproceedings{lin2017feature,
  title     = {Feature Pyramid Networks for Object Detection},
  author    = {Lin, Tsung-Yi and Doll{\'a}r, Piotr and Girshick, Ross and He, Kaiming and Hariharan, Bharath and Belongie, Serge},
  booktitle = {Proceedings of the IEEE Conference on Computer Vision and Pattern Recognition (CVPR)},
  pages     = {936--944},
  year      = {2017}
}

@article{hu2021lora,
  title={Lora: Low-rank adaptation of large language models. arXiv 2021},
  author={Hu, Edward J and Shen, Yelong and Wallis, Phillip and Allen-Zhu, Zeyuan and Li, Yuanzhi and Wang, Shean and Wang, Lu and Chen, Weizhu},
  journal={arXiv preprint arXiv:2106.09685},
  volume={10},
  year={2021}
}

@inproceedings{fini2025multimodal,
  title={Multimodal autoregressive pre-training of large vision encoders},
  author={Fini, Enrico and Shukor, Mustafa and Li, Xiujun and Dufter, Philipp and Klein, Michal and Haldimann, David and Aitharaju, Sai and da Costa, Victor G Turrisi and B{\'e}thune, Louis and Gan, Zhe and others},
  booktitle={Proceedings of the Computer Vision and Pattern Recognition Conference},
  pages={9641--9654},
  year={2025}
}

@inproceedings{kamann2020benchmarking,
  title     = {Benchmarking the Robustness of Semantic Segmentation Models},
  author    = {Kamann, Christoph and Rother, Carsten},
  booktitle = {Proceedings of the IEEE/CVF Conference on Computer Vision and Pattern Recognition (CVPR)},
  year      = {2020}
}

@inproceedings{kerssies2024benchmark,
  title={How to Benchmark Vision Foundation Models for Semantic Segmentation?},
  author={Kerssies, Tommie and De Geus, Daan and Dubbelman, Gijs},
  booktitle={Proceedings of the IEEE/CVF Conference on Computer Vision and Pattern Recognition},
  pages={1162--1171},
  year={2024}
}

@article{chen2022pali,
  title={Pali: A jointly-scaled multilingual language-image model},
  author={Chen, Xi and Wang, Xiao and Changpinyo, Soravit and Piergiovanni, Anthony J and Padlewski, Piotr and Salz, Daniel and Goodman, Sebastian and Grycner, Adam and Mustafa, Basil and Beyer, Lucas and others},
  journal={arXiv preprint arXiv:2209.06794},
  year={2022}
}

@article{hou2024lmdeeplabv3plus,
  title     = {LM-DeeplabV3+: A Lightweight Image Segmentation Algorithm Based on Multi-Scale Feature Interaction},
  author    = {Hou, Xinyu and Chen, Peng and Gu, Haishuo},
  journal   = {Applied Sciences},
  volume    = {14},
  number    = {4},
  pages     = {1558},
  year      = {2024},
  publisher = {MDPI},
  doi       = {10.3390/app14041558}
}

@inproceedings{goyal2017making,
  title={Making the v in vqa matter: Elevating the role of image understanding in visual question answering},
  author={Goyal, Yash and Khot, Tejas and Summers-Stay, Douglas and Batra, Dhruv and Parikh, Devi},
  booktitle={Proceedings of the IEEE conference on computer vision and pattern recognition},
  pages={6904--6913},
  year={2017}
}

@article{krishna2017visual,
  title={Visual genome: Connecting language and vision using crowdsourced dense image annotations},
  author={Krishna, Ranjay and Zhu, Yuke and Groth, Oliver and Johnson, Justin and Hata, Kenji and Kravitz, Joshua and Chen, Stephanie and Kalantidis, Yannis and Li, Li-Jia and Shamma, David A and others},
  journal={International journal of computer vision},
  volume={123},
  number={1},
  pages={32--73},
  year={2017},
  publisher={Springer}
}

@article{zhou2024tinyllava,
  title={Tinyllava: A framework of small-scale large multimodal models},
  author={Zhou, Baichuan and Hu, Ying and Weng, Xi and Jia, Junlong and Luo, Jie and Liu, Xien and Wu, Ji and Huang, Lei},
  journal={arXiv preprint arXiv:2402.14289},
  year={2024}
}

@inproceedings{hudson2019gqa,
  title={Gqa: A new dataset for real-world visual reasoning and compositional question answering},
  author={Hudson, Drew A and Manning, Christopher D},
  booktitle={Proceedings of the IEEE/CVF conference on computer vision and pattern recognition},
  pages={6700--6709},
  year={2019}
}

@inproceedings{singh2019towards,
  title={Towards vqa models that can read},
  author={Singh, Amanpreet and Natarajan, Vivek and Shah, Meet and Jiang, Yu and Chen, Xinlei and Batra, Dhruv and Parikh, Devi and Rohrbach, Marcus},
  booktitle={Proceedings of the IEEE/CVF conference on computer vision and pattern recognition},
  pages={8317--8326},
  year={2019}
}

@article{sun2023eva,
  title        = {EVA-CLIP: Improved Training Techniques for CLIP at Scale},
  author       = {Sun, Quan and Fang, Yuxin and Wu, Ledell and Wang, Xinlong and Cao, Yue},
  journal      = {arXiv preprint arXiv:2303.15389},
  year         = {2023}
}

@inproceedings{pan2023stitchable,
  title        = {Stitchable Neural Networks},
  author       = {Pan, Zizheng and Zhuang, Bohan and He, Haoyu and Liu, Jing and Cai, Jianfei},
  booktitle    = {Proceedings of the IEEE/CVF Conference on Computer Vision and Pattern Recognition (CVPR)},
  pages        = {16041--16050},
  year         = {2023}
}

@inproceedings{aaai2025tasklossmatch,
  title        = {How not to Stitch Representations to Measure Similarity: Task Loss Matching versus Direct Matching},
  author       = {Collins, Katherine M and Bhatt, Umang and Weller, Adrian},
  booktitle    = {Proceedings of the AAAI Conference on Artificial Intelligence},
  year         = {2025}
}

@article{fang2023data,
  title        = {Data Filtering Networks},
  author       = {Fang, Alex and Jose, Albin Madappally and Jain, Amit and Schmidt, Ludwig and Toshev, Alexander and Shankar, Vaishaal},
  journal      = {arXiv preprint arXiv:2309.17425},
  year         = {2023}
}

@article{krizhevsky2009learning,
  title={Learning Multiple Layers of Features from Tiny Images},
  author={Krizhevsky, Alex and Hinton, Geoffrey and others},
  year={2009},
  publisher={Citeseer}
}
}

\clearpage
\setcounter{page}{1}
\maketitlesupplementary

% Define custom colors to match the image
\definecolor{cvprblue}{RGB}{0, 0, 255}
\definecolor{cvprorange}{RGB}{255, 165, 0}

\appendix
\noindent{We provide details omitted in the paper.}

\begin{itemize}
    \item \cref{sec_sup:exp_detail}: {Experiment and dataset details}
    \item \cref{sec_sup:detail_results}: {Detailed results}
\end{itemize}

\section{Experiment and Dataset Details}
\label{sec_sup:exp_detail}

\subsection{Vision Foundation Models (VFMs)}
We use the following VFMs in our experiments.
\begin{itemize}
    \item \textbf{DINOv2-L}~\cite{oquab2023dinov2}: A ViT-L/14 encoder trained with self-supervised DINOv2 on a curated collection of $\sim$142M internet images, using a teacher–student distillation objective that encourages consistency across multi-crop image views without labels. In our experiments, we feed $336 \times 336$ inputs with patch size $14$.

    \item \textbf{CLIP-L}~\cite{radford2021learning}: A ViT-L/14 vision encoder from CLIP, trained jointly with a text encoder on $\sim$400M image–text pairs from the web using a contrastive objective that pulls matched image–caption pairs together and pushes mismatched pairs apart in a shared embedding space. We use the vision tower with $336 \times 336$ inputs and patch size $14$.

    \item \textbf{SigLIP2-L}~\cite{siglip2}: A ViT-L/16 multilingual vision–language encoder trained on a diverse mixture of web image–text data with a sigmoid-based image–text matching loss, augmented with captioning-style pretraining, self-distillation, masked prediction, and online data curation to improve dense features and localization. We use $384 \times 384$ inputs and patch size $16$.

    \item \textbf{DINOv3-L}~\cite{simeoni2025dinov3}: A ViT-L/16 self-supervised encoder from the DINOv3 family, trained at large scale on a multi-domain image collection with a distillation-style objective and Gram anchoring to stabilize dense feature maps, yielding strong frozen representations for both global and dense tasks. In our experiments, we use $384 \times 384$ inputs and patch size $16$.
\end{itemize}

\subsection{Fine-Grained Classification}

\subsubsection{Dataset Details}
\paragraph{fMoW~\cite{christie2018functional}}
The Functional Map of the World (fMoW) dataset is a commonly used and challenging dataset for VFM evaluation. It is a collection of high-quality remote-sensing satellite images collected worldwide, annotated with 62 land-use and functional categories (e.g., airfield, port, hospital). We use the fMoW-RGB variant, which provides pan-sharpened RGB crops and associated metadata. Since the original dataset is highly imbalanced, we construct a class-balanced subset with 230 training and 26 test images per class (14{,}260 train / 1{,}612 test) so that our study can focus on model stitching itself rather than confounding effects from label imbalance, which would complicate the interpretation of stitching behavior.

\paragraph{iNaturalist-Subset~\cite{van2018inaturalist}}
The iNaturalist 2021 dataset family has become a standard testbed for assessing VFM performance on fine-grained, real-world biodiversity recognition. It is a large-scale and heavily imbalanced species classification benchmark that reflects naturally occurring long-tailed distributions. To obtain a controlled yet challenging setting for analyzing stitching, we sample 106 species from three visually similar Lepidoptera families (Sphingidae, Pieridae, Pyralidae) and rebalance the data with 200 training and 50 test images per class (21{,}200 train / 5{,}300 test). This balanced construction removes imbalance as an additional variable, enabling a cleaner analysis of how stitching choices impact performance.

\paragraph{FGVC-Aircraft~\cite{maji2013fgvc}}
FGVC-Aircraft is widely adopted as a canonical fine-grained classification benchmark for VFMs. The version we use contains 102 aircraft models and approximately 10k images, with each model appearing in around 100 images. We follow the standard split with 6.6k training and 3.3k test images, which requires distinguishing subtle variations in shape, livery, and viewpoint.

\subsubsection{Training Details}
For all fine-grained classification experiments, we adopt linear probing: a single linear classifier is trained on pooled image features while all VFM parameters remain frozen. 

For \emph{layer feature matching}, we first extract intermediate features from both source and target VFMs and train the stitch MLP purely on these features; no additional VFM forward passes are required during this phase. For \emph{final feature matching}, we extract features from the target model and train only the stitch layer.

To isolate the effect of stitching, we do not apply any data augmentation. We use the AdamW optimizer in all experiments, train for up to 100 epochs with early stopping (patience of 5 epochs), and tune the learning rate in $\{0.001, 0.005, 0.01\}$. For layer feature matching, we use a batch size of 256; all other configurations use a batch size of 128. Training is performed with automatic mixed precision using \texttt{bfloat16}.

\subsubsection{Semantic Segmentation}
\subsubsection{Dataset Details}

\paragraph{ADE20K~\cite{zhou2017scene,zhou2019semantic}}
ADE20K is a scene-centric semantic segmentation dataset with pixel-level annotations for 150 object and stuff categories across diverse indoor and outdoor environments. We adopt the canonical split with 20{,}210 training images and 3{,}000 held-out test images. The dense annotations cover scenes, objects, and parts, making ADE20K a challenging benchmark for evaluating dense prediction performance.

\subsubsection{Training Details}
For semantic segmentation, we train a linear layer to predict per-pixel class logits for each patch token. The linear layer produces a low-resolution logit map (e.g., $24 \times 24$ for a model with patch size 14 and $336 \times 336$ input), which is then bilinearly upsampled to the original image resolution to obtain the final segmentation map.

For layer feature matching and final feature matching, we reuse the optimization and hyperparameter settings described for classification (optimizer, learning rates, batch sizes, early stopping, and mixed precision). All remaining experimental details for segmentation follow~\cite{kerssies2024benchmark}.

\subsection{VFM Stitch Tree for Multimodal LLM}
\subsubsection{Dataset Details}
We use the LLaVA-1.5 training data prepared by TinyLLaVA.\footnote{We use the data preparation pipeline from TinyLLaVA: \url{https://tinyllava-factory.readthedocs.io/en/latest/Prepare\%20Datasets.html}} 
LLaVA-1.5-Pretrain (PT) consists of 558k image–caption pairs~\cite{liu2024improved}, while LLaVA-1.5-SFT comprises 665k visual instruction-tuning conversations that combine academic-style VQA~\cite{goyal2017making, singh2019towards, krishna2017visual, hudson2019gqa} samples with instruction-tuning data from LLaVA-Instruct~\cite{liu2023visual}. In our preliminary study, we evaluate on VQA-v2~\cite{goyal2017making} and MME~\cite{fu2023mme}.

\subsubsection{Training Details}
We jointly use LLaVA-1.5-Pretrain and LLaVA-1.5-SFT to train the stitch layer for one epoch with a learning rate of 0.001 and batch size 64. All remaining hyperparameters follow the TinyLLaVA recipe~\cite{zhou2024tinyllava}. We adopt Qwen2.5-3B~\cite{bai2023qwen} as the LLM and employ an interleaved mixture-of-features (MoF)~\cite{tong2024eyes} strategy when combining visual tokens from CLIP and DINOv2.

\section{Detailed and Additional Results}
\label{sec_sup:detail_results}
\subsection{Self-Stitch}
\begin{figure}
    \centering
    \includegraphics[width=1\linewidth]{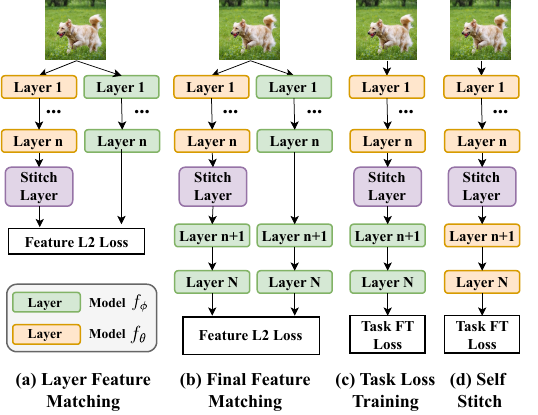}    \caption{Model stitching training strategies: (a) Layer feature matching trains the stitch layer to match features between the source and target models at the stitch position. (b) Final feature matching trains the stitch layer so that the stitched model matches the target model’s final features. (c) Task loss training optimizes the stitch layer with the downstream task objective. (d) Self-Stitch inserts the stitch layer into the source-only and target-only models at the same stitch position. Across all strategies, the stitch layer is the only trainable component; the source and target models are kept frozen. }
    \label{fig_sup:self}
\end{figure}

To rigorously test whether gains only stem from added capacity,  we introduce a Self-Stitch baseline: inserting the identical stitch layer into the source-only and target-only models at the same stitch positions. \cref{fig_sup:self} illustrates the difference between model stitching and the self-stitching baseline.

\subsection{Layer and Final Feature Matching}

We present a detailed comparison between Layer and Final Feature Matching in \cref{tab_sup:feature_matching}. We observe that Final Feature Matching consistently outperforms Layer Feature Matching across all configurations. Notably, in the DINOv2 $\to$ SigLIP2 setting, the stitched model surpasses the performance of both constituent models. Our results further indicate that stitching a lower-performing source model to a higher-performing target is more effective than the reverse; DINOv2 $\to$ SigLIP2 consistently achieves higher accuracy than SigLIP2 $\to$ DINOv2, regardless of the matching strategy employed.

\begin{table}[t]
    \centering

    \resizebox{\linewidth}{!}{%
        \setlength{\tabcolsep}{3.5pt}
        % User requested column spec with vertical lines
        \begin{tabular}{l|c|c|c|c|c|c}
            \toprule
             & \multicolumn{3}{c}{\textcolor{cvprblue}{DINOv2}} & \multicolumn{3}{c}{\textcolor{cvprorange}{SigLIP2}} \\
             \midrule
            Linear Probing & \multicolumn{3}{c}{46.7} & \multicolumn{3}{c}{53.5} \\
            \midrule
            Layer & 2 & 6 & 10 & 14 & 18 & 22 \\
            \midrule
            \multicolumn{7}{c}{\textbf{Layer Feature Matching}} \\
            \midrule
            DINOv2 $\to$ SigLIP2 & \textcolor{cvprblue}{49.4} & \textcolor{cvprblue}{48.2} & \textcolor{cvprblue}{51.0} & \textcolor{cvprblue}{48.5} & 45.3 & \textcolor{cvprblue}{48.2} \\
            SigLIP2 $\to$ DINOv2 & 29.9 & 32.4 & 26.9 & 42.4 & 43.7 & \textcolor{cvprblue}{47.2} \\
            \midrule
            \multicolumn{7}{c}{\textbf{Final Feature Matching}} \\
            \midrule
            DINOv2 $\to$ SigLIP2 & \textcolor{cvprblue}{51.4} & \textcolor{cvprblue}{53.1} & \textcolor{cvprorange}{56.1} & \textcolor{cvprblue}{51.0} & \textcolor{cvprorange}{55.7} & \textcolor{cvprorange}{54.7} \\
            SigLIP2 $\to$ DINOv2 & 46.4 & \textcolor{cvprblue}{47.1} & 43.3 & \textcolor{cvprblue}{47.3} & \textcolor{cvprblue}{48.8} & \textcolor{cvprblue}{50.8} \\
            \bottomrule
        \end{tabular}%
    }
        \caption{\textbf{Comparison of Layer and Final Feature Matching.} We report linear probing accuracy for different stitching strategies. \textcolor{cvprorange}{Orange} indicates performance surpassing the SigLIP2 baseline, while \textcolor{cvprblue}{Blue} indicates performance surpassing the DINOv2 baseline.}
    \label{tab_sup:feature_matching}
\end{table}

\subsection{Stitching Between Different VFMs}

To verify the generality of our findings, we extend our analysis beyond DINOv2 and SigLIP2 to include the widely adopted CLIP and the recently released DINOv3. Detailed comparisons between Stitched models, Self-Stitch baselines, and Linear Probing are provided in \cref{tab_sup:cd2,tab_sup:d2s,tab_sup:sc,tab_sup:d3s,tab_sup:d2d3}. Consistent with our previous results, stitched models generally outperform their corresponding self-stitch baselines across most layers. The primary exception arises when CLIP is utilized as the source model; we discuss the potential causes for this phenomenon in the main paper.

\begin{table}[h]
    \centering

    \resizebox{\linewidth}{!}{%
    \setlength{\tabcolsep}{12pt} % Adjust spacing to match wide look of screenshot
    \begin{tabular}{lcccc}
        \toprule
         & \multicolumn{2}{c}{DINOv2} & \multicolumn{2}{c}{CLIP} \\
         \midrule
        Linear Probing & \multicolumn{2}{c}{46.7} & \multicolumn{2}{c}{46.4} \\
        \midrule
        Layer & 6 & 14 & 18 & 22 \\
        \midrule
        DINOv2 $\to$ DINOv2 & 41.5 & 59.7 & 68.2 & 69.9 \\
        CLIP $\to$ CLIP     & 48.5 & 63.1 & 59.6 & 60.7 \\
        \midrule
        CLIP $\to$ DINOv2   & 53.1 & 65.8 & 63.3 & 64.0 \\
        DINOv2 $\to$ CLIP   & \textbf{54.1} & \textbf{68.1} & \textbf{72.0} & \textbf{71.9} \\
        \bottomrule
    \end{tabular}
    }
        \caption{Stitched Model vs Self-Stitch vs Linear Probing: DINOv2 and CLIP.}
        \label{tab_sup:cd2}
\end{table}

\begin{table}[h]
    \centering

    \resizebox{\linewidth}{!}{%
    \setlength{\tabcolsep}{12pt}
    \begin{tabular}{lcccc}
        \toprule
         & \multicolumn{2}{c}{DINOv2} & \multicolumn{2}{c}{SigLIP2} \\
        \midrule
        Linear Probing & \multicolumn{2}{c}{46.7} & \multicolumn{2}{c}{53.5} \\
        \midrule
        Layer & 6 & 14 & 18 & 22 \\
        \midrule
        DINOv2 $\to$ DINOv2   & 41.5 & 59.7 & 68.2 & 69.9 \\
        SigLIP2 $\to$ SigLIP2 & 50.5 & 62.0 & 69.4 & 68.9 \\
        \midrule
        SigLIP2 $\to$ DINOv2  & 53.8 & \textbf{69.6} & 70.4 & \textbf{72.2} \\
        DINOv2 $\to$ SigLIP2  & \textbf{55.8} & 68.0 & \textbf{72.0} & 71.8 \\
        \bottomrule
    \end{tabular}
    }
        \caption{Stitched Model vs Self-Stitch vs Linear Probing: DINOv2 and SigLIP2.}
        \label{tab_sup:d2s}
\end{table}

\begin{table}[h]
    \centering

    \resizebox{\linewidth}{!}{%
    \setlength{\tabcolsep}{12pt}
    \begin{tabular}{lcccc}
        \toprule
         & \multicolumn{2}{c}{SigLIP2} & \multicolumn{2}{c}{CLIP} \\
        \midrule
        Linear Probing & \multicolumn{2}{c}{53.5} & \multicolumn{2}{c}{46.4} \\
        \midrule
        Layer & 6 & 14 & 18 & 22 \\
        \midrule
        SigLIP2 $\to$ SigLIP2 & 50.5 & 62.0 & 69.4 & 68.9 \\
        CLIP $\to$ CLIP       & 48.5 & 63.1 & 59.6 & 60.7 \\
        \midrule
        CLIP $\to$ SigLIP2    & 48.3 & 68.9 & 65.4 & 62.9 \\
        SigLIP2 $\to$ CLIP    & \textbf{59.8} & \textbf{70.7} & \textbf{73.2} & \textbf{71.9} \\
        \bottomrule
    \end{tabular}
    }
        \caption{Stitched Model vs Self-Stitch vs Linear Probing: SigLIP2 and CLIP.}
        \label{tab_sup:sc}
\end{table}

\begin{table}[h]
    \centering

    \resizebox{\linewidth}{!}{%
    \setlength{\tabcolsep}{12pt}
    \begin{tabular}{lcccc}
        \toprule
         & \multicolumn{2}{c}{DINOv3} & \multicolumn{2}{c}{SigLIP2} \\
        \midrule
        Linear Probing & \multicolumn{2}{c}{50.0} & \multicolumn{2}{c}{53.5} \\
        \midrule
        Layer & 6 & 14 & 18 & 22 \\
        \midrule
        DINOv3 $\to$ DINOv3   & 44.7 & 63.9 & 66.9 & 69.0 \\
        SigLIP2 $\to$ SigLIP2 & 50.5 & 62.0 & 69.4 & 68.9 \\
        \midrule
        SigLIP2 $\to$ DINOv3  & \textbf{58.6} & 67.0 & 69.4 & 69.3 \\
        DINOv3 $\to$ SigLIP2  & 45.0 & \textbf{69.4} & \textbf{70.5} & \textbf{72.4} \\
        \bottomrule
    \end{tabular}
    }
        \caption{Stitched Model vs Self-Stitch vs Linear Probing: DINOv3 and SigLIP2.}
        \label{tab_sup:d3s}
\end{table}

\begin{table}[h]
    \centering

    \resizebox{\linewidth}{!}{%
    \setlength{\tabcolsep}{12pt}
    \begin{tabular}{lcccc}
        \toprule
         & \multicolumn{2}{c}{DINOv2} & \multicolumn{2}{c}{DINOv3} \\
        \midrule
        Linear Probing & \multicolumn{2}{c}{46.7} & \multicolumn{2}{c}{50.0} \\
        \midrule
        Layer & 6 & 14 & 18 & 22 \\
        \midrule
        DINOv2 $\to$ DINOv2 & 41.5 & 59.7 & 68.2 & 69.9 \\
        DINOv3 $\to$ DINOv3 & 44.7 & 63.9 & 66.9 & 69.0 \\
        \midrule
        DINOv3 $\to$ DINOv2 & 43.8 & \textbf{67.2} & 67.9 & \textbf{72.0} \\
        DINOv2 $\to$ DINOv3 & \textbf{57.8} & 65.0 & \textbf{69.2} & 70.3 \\
        \bottomrule
    \end{tabular}
    }
        \caption{Stitched Model vs Self-Stitch vs Linear Probing: DINOv2 and DINOv3.}
        \label{tab_sup:d2d3}
\end{table}

\subsection{Prediction Analysis}

\newcommand{\Rcell}{\cellcolor{green!20}R}
\newcommand{\Wcell}{\cellcolor{red!20}W}

\begin{table*}[t]
\centering
\setlength{\tabcolsep}{4pt}
\renewcommand{\arraystretch}{1.1}
\begin{tabular}{cc}
\multicolumn{2}{c}{\textbf{fMoW}} \\
\midrule
{
\begin{tabular}{lccccccc}
 & \multicolumn{6}{c}{Scenario} & Acc. \\
\midrule
DINOv2$\rightarrow$DINOv2   & \Wcell & \Rcell & \Wcell & \Rcell & \Rcell & \Wcell & 69.9 \\
SigLIP2$\rightarrow$SigLIP2 & \Rcell & \Wcell & \Wcell & \Rcell & \Wcell & \Rcell & 68.9 \\
\textbf{SigLIP2$\rightarrow$DINOv2}  & \Rcell & \Rcell & \Rcell & \Wcell & \Wcell & \Wcell & 72.2 \\
\midrule
Count                       & 116    & 78     & 47     & 40     & 86     & 33     &      \\
\end{tabular}
}
&
{
\begin{tabular}{lccccccc}
  & \multicolumn{6}{c}{Scenario}  & Acc. \\
\midrule
DINOv2$\rightarrow$DINOv2   & \Wcell & \Rcell & \Wcell & \Rcell & \Rcell & \Wcell & 69.9 \\
SigLIP2$\rightarrow$SigLIP2 & \Rcell & \Wcell & \Wcell & \Rcell & \Wcell & \Rcell & 68.9 \\
\textbf{DINOv2$\rightarrow$SigLIP2}  & \Rcell & \Rcell & \Rcell & \Wcell & \Wcell & \Wcell & 71.8 \\
\midrule
Count                       & 77     & 116    & 54     & 52     & 48     & 72     &      \\
\end{tabular}
}
\\[8pt]

\multicolumn{2}{c}{\textbf{iNaturalist}} \\
\midrule
{
\begin{tabular}{lccccccc}
 & \multicolumn{6}{c}{Scenario} & Acc. \\
\midrule
DINOv2$\rightarrow$DINOv2   & \Wcell & \Rcell & \Wcell & \Rcell & \Rcell & \Wcell & 91.2 \\
SigLIP2$\rightarrow$SigLIP2 & \Rcell & \Wcell & \Wcell & \Rcell & \Wcell & \Rcell & 87.3 \\
\textbf{SigLIP2$\rightarrow$DINOv2}  & \Rcell & \Rcell & \Rcell & \Wcell & \Wcell & \Wcell & 91.9 \\
\midrule
Count                       & 138    & 292    & 75     & 68     & 116    & 53     &      \\
\end{tabular}
}
&
{
\begin{tabular}{lccccccc}
 & \multicolumn{6}{c}{Scenario}  & Acc. \\
\midrule
DINOv2$\rightarrow$DINOv2   & \Wcell & \Rcell & \Wcell & \Rcell & \Rcell & \Wcell & 91.2 \\
SigLIP2$\rightarrow$SigLIP2 & \Rcell & \Wcell & \Wcell & \Rcell & \Wcell & \Rcell & 87.3 \\
\textbf{DINOv2$\rightarrow$SigLIP2}  & \Rcell & \Rcell & \Rcell & \Wcell & \Wcell & \Wcell & 92.8 \\
\midrule
Count                       & 127    & 331    & 80     & 49     & 77     & 64     &      \\
\end{tabular}
}
\\
\end{tabular}
\caption{
Analysis of Predictions: \colorbox{green!20}{R} and \colorbox{red!20}{W} denote Right and Wrong predictions for each model, with counts in the last row and accuracies in the last column for fMoW and iNaturalist. Stitched models fall into complementarity scenarios (stitched is correct while at least one self-stitched baseline is wrong) far more often than interference ones (stitched is wrong while at least one self-stitched baseline is correct). When the two self-stitched models disagree, the stitched model tends to follow the source model on fMoW but follows the stronger model DINOv2 on iNaturalist.
}

\label{tab_sup:pred}
\end{table*}

To better understand how knowledge fusion affects prediction behavior, we compare each stitched model (DINOv2$\rightarrow$SigLIP2 and SigLIP2$\rightarrow$DINOv2) against the two self-stitched baselines (DINOv2$\rightarrow$DINOv2 and SigLIP2$\rightarrow$SigLIP2) on fMoW and iNaturalist. We discard trivial cases where all three models are correct or all three are wrong, and partition the remaining examples into the scenarios in \cref{tab_sup:pred}:

\begin{enumerate}
    \item \textbf{Preserve (Cols.\ 1--2).} At least one self-stitched model is correct and the stitched model is also correct, preserving the correct prediction.
    \item \textbf{Rescue (Col.\ 3).} Both self-stitched models are wrong but the stitched model is correct.
    \item \textbf{Interference (Cols.\ 4--6).} At least one self-stitched model is correct but the stitched model becomes wrong.
\end{enumerate}

Across both datasets and stitch directions, the total count of \emph{preserve} and \emph{rescue} scenarios clearly exceeds that of \emph{interference} scenarios, indicating that the stitched models benefit more from fusing complementary signals than they suffer from conflicts between the two backbones.

We further ask whose behavior the stitched model tends to follow when the two self-stitched models disagree. On fMoW, where the two self-stitched accuracies are similar, the stitched models are more likely to match the \emph{source} model: for SigLIP2$\rightarrow$DINOv2, the number of cases where the stitched prediction agrees with SigLIP2$\rightarrow$SigLIP2 substantially exceeds the cases where it agrees with DINOv2$\rightarrow$DINOv2, and an analogous trend holds for DINOv2$\rightarrow$SigLIP2. In contrast, on iNaturalist, where DINOv2$\rightarrow$DINOv2 is the stronger self-stitched model, both stitch directions tend to align with the stronger model’s predictions regardless of whether it is used as source or target.

Overall, these analyses support our interpretation that stitching primarily acts as a knowledge-fusion mechanism: it preserves or rescues correct predictions from at least one backbone much more often than it disrupts them, and when there is a strong/weak asymmetry, the stitched model gravitates toward the stronger expert’s decisions.

\subsection{VFM Stitch Tree (VST) for Multimodal LLM (MLLM)}

\subsubsection{Computation Comparison}

\Cref{tab_sup:vst_cost} compares the computational cost of running all VFMs independently (Full) versus our VST method at various stitch positions. We illustrate the efficiency gains using Cambrian-1 with \textbf{4} VFMs~\cite{tong2024cambrian}. Assuming a standard 24-layer ViT-L architecture~\cite{siglip2}, and noting that MLLMs typically extract features from the second last layer (layer 23)~\cite{liu2024improved}, we base our calculations on a 23-layer depth. While the ``Full'' setting incurs a 300\% computational overhead compared to a single VFM, VST-14 significantly reduces this burden. By sharing the first 14 layers and maintaining specialized branches only from layer 15 onwards, VST-14 requires processing only $3 \times (23-14) = 27$ additional layers. This results in an overhead of just $27 / 23 \approx 117\%$, demonstrating a substantial efficiency improvement over the independent baseline.

\begin{table}[h]
    \centering

    \resizebox{\linewidth}{!}{%
        \setlength{\tabcolsep}{8pt} % Adjust spacing for a clean look
        \begin{tabular}{ccccc}
            \toprule
            Number of VFMs & Full & VST-6 & VST-14 & VST-22 \\
            \midrule
            2 & 100\% & 74\% & 39\% & 4\% \\
            3 & 200\% & 148\% & 78\% & 9\% \\
            4 & 300\% & 222\% & 117\% & 13\% \\
            5 & 400\% & 296\% & 156\% & 17\% \\
            \bottomrule
        \end{tabular}%
    }
        \caption{\textbf{Comparison of Additional Computation Cost.} We report the additional increase in computation compared to a single VFM. ``Full'' denotes running all VFMs independently. ``VST-$n$'' denotes a shared backbone up to the stitch position $n$, where layers $1$ through $n$ are shared and subsequent layers are executed independently. For example, VST-14 shares the first 14 layers and maintains specialized branches from layer 15 onwards.}
    \label{tab_sup:vst_cost}
\end{table}

\subsection{VST as an Accuracy-Efficiency Knob}
\label{sec:appendix_calc}
\begin{table*}[t]
    \centering

    \resizebox{\linewidth}{!}{%
        \setlength{\tabcolsep}{8pt}
        \begin{tabular}{l|ccc|cc|cc}
            \toprule
            \textbf{Model Configuration} & \multicolumn{3}{c|}{\textbf{VQAv2}} & \multicolumn{2}{c|}{\textbf{MME}} & \multicolumn{2}{c}{\textbf{Efficiency Metrics}} \\
             & Yes/No & Number & Other & Percep. & Cogn. & \textbf{Avg Gain \%} & \textbf{Add. Cost} \\
            \midrule
            
            \multicolumn{8}{l}{\textit{\textbf{1. Define the Upper Bound (Denominator)}}} \\
            (A) CLIP Baseline & 91.75 & 58.74 & 69.00 & 1418.5 & 277.1 & - & 0\% \\
            (B) Full (CLIP + DINOv2) & 92.72 & 61.64 & 70.30 & 1460.3 & 311.8 & - & 100\% \\
            % The Denominator Row (Orange)
            \rowcolor{denomOrange}
            \textbf{(C) Max Gain ($\Delta_{\text{max}} = B - A$)} & \textbf{0.97} & \textbf{2.90} & \textbf{1.30} & \textbf{41.8} & \textbf{34.7} & \textbf{(Denom.)} & \textbf{-} \\
            
            \midrule
            \multicolumn{8}{l}{\textit{\textbf{2. VST-22: The Lightweight Knob (High Efficiency)}}} \\
            (D) VST-22 (Ours) & 92.12 & 59.21 & 69.15 & 1451.6 & 305.7 & - & \textbf{4.3\%} \\
            % The Numerator Row (Pink)
            \rowcolor{numPink}
            \textbf{(E) VST Gain ($\Delta_{\text{VST}} = D - A$)} & \textbf{0.37} & \textbf{0.47} & \textbf{0.15} & \textbf{33.1} & \textbf{28.6} & \textbf{(Num.)} & \textbf{-} \\
            \textbf{Normalized Gain ($\% = E / C$)} & 38.1\% & 16.2\% & 11.5\% & 79.1\% & 82.5\% & \textbf{45.5\%} & \textbf{4.3\%} \\

            \midrule
            \multicolumn{8}{l}{\textit{\textbf{3. VST-14: The Balanced Knob (High Performance)}}} \\
            (F) VST-14 (Ours) & 92.54 & 60.69 & 69.88 & 1474.4 & 301.8 & - & 39.0\% \\
            % The Numerator Row (Pink)
            \rowcolor{numPink}
            \textbf{(G) VST Gain ($\Delta_{\text{VST}} = F - A$)} & \textbf{0.79} & \textbf{1.95} & \textbf{0.88} & \textbf{55.9} & \textbf{24.7} & \textbf{(Num.)} & \textbf{-} \\
            \textbf{Normalized Gain ($\% = G / C$)} & 81.4\% & 67.2\% & 67.7\% & 133.7\% & 71.1\% & \textbf{84.2\%} & 39.0\% \\
            \bottomrule
        \end{tabular}%
    }
        \caption{\textbf{Detailed Calculation of Normalized Gain.} Comparison of the ``Lightweight'' VST-22 and the ``Balanced'' VST-14. The \colorbox{denomOrange}{Orange Row} represents the  gain ($\Delta_{\text{max}}$) achieved by the computationally expensive Naive ensemble (100\% Cost). The \colorbox{numPink}{Pink Rows} represent the actual gain ($\Delta_{\text{VST}}$) achieved by our method. The \textbf{Normalized Gain} is calculated as $\Delta_{\text{VST}} / \Delta_{\text{max}}$. Note that VST-22 achieves nearly half the total gain (45.5\%) with negligible cost (4.3\%).}
    \label{tab_sup:gain_calculation}
\end{table*}
We introduce the concept of ``Normalized Gain'' to quantify the effectiveness of VST relative to running all VFMs independently. \Cref{tab_sup:gain_calculation} provides the step-by-step derivation of these values for both VST-22 and VST-14.

We define the Normalized Gain as the ratio between the performance improvement achieved by VST ($\Delta_{\text{VST}}$) and the maximum improvement achieved by running all VFMs independently ($\Delta_{\text{max}}$):
\begin{equation}
    \text{Normalized Gain (\%)} = \frac{\Delta_{\text{VST}}}{\Delta_{\text{max}}} = \frac{\text{Perf}_{\text{VST}} - \text{Perf}_{\text{CLIP}}}{\text{Perf}_{\text{Full}} - \text{Perf}_{\text{CLIP}}}
\end{equation}

As illustrated in \cref{tab_sup:gain_calculation}:
\begin{itemize}
    \item The \colorbox{denomOrange}{Orange Row} represents the \textbf{Denominator} ($\Delta_{\text{max}}$): the total performance gap between the single-model baseline and the computationally expensive naive full running (100\% additional cost).
    \item The \colorbox{numPink}{Pink Rows} represent the \textbf{Numerator} ($\Delta_{\text{VST}}$): the actual gain realized by our VST method.
\end{itemize}

The results highlight the versatility of VST as a compute-aware knob. \textbf{VST-22} serves as an ultra-lightweight option, recovering \textbf{45.5\%} of the total possible gain while incurring a negligible \textbf{4.3\%} increase in backbone cost. Conversely, \textbf{VST-14} acts as a balanced high-performance option, recovering \textbf{84.2\%} of the total gain for only \textbf{39\%} of the additional cost. This granular control allows practitioners to maximize model utility within specific computational budgets.

Consequently, we view VST as a vital \textit{accuracy--efficiency knob} for architecture design. Without VST, practitioners face a rigid binary choice: either deploy an entire additional VFM (yielding higher performance but incurring \textbf{100\%} extra backbone cost) or deploy none (yielding no gain). VST transforms this discrete step function into a continuous spectrum. By serving as a compute-aware knob, our method effectively interpolates between these extremes, enabling controllable performance--efficiency trade-offs that can be dynamically tuned to meet strict real-world deployment constraints.

\end{document}